\documentclass[10pt,letterpaper]{article}
\usepackage[top=0.85in,left=2.75in,footskip=0.75in]{geometry}

\usepackage{amsmath,amssymb}

\usepackage{changepage}

\usepackage[utf8x]{inputenc}

\usepackage{textcomp,marvosym}

\usepackage{cite}
\usepackage{cite}
\usepackage{amsmath,amssymb,amsfonts}
\usepackage{algorithmic}
\usepackage{graphicx}
\usepackage{textcomp}
\usepackage{microtype}
\usepackage{booktabs}
\usepackage{adjustbox}
\usepackage[utf8x]{inputenc}

\usepackage{nccmath}
\usepackage{makecell}

\usepackage{nameref,hyperref}
\usepackage{subfigure}
\usepackage[right]{lineno}

\usepackage{graphicx}
\usepackage{textcomp}
\usepackage{microtype}
\usepackage{booktabs}
\usepackage{microtype}
\DisableLigatures[f]{encoding = *, family = * }

\usepackage{array}
\usepackage{caption}
\usepackage{url}  
\usepackage{array,multirow} 
\usepackage{ragged2e,multicol}

\usepackage{cite} 
 
\usepackage{textcomp,marvosym}

\usepackage{siunitx}
\usepackage{tabularx}
\usepackage{algorithmic}
\usepackage{easyReview}

\newcolumntype{+}{!{\vrule width 2pt}}

\newlength\savedwidth

\raggedright
\usepackage[aboveskip=1pt,labelfont=bf,labelsep=period,justification=raggedright,singlelinecheck=off]{caption}
\renewcommand{\figurename}{Fig}

\makeatletter
\renewcommand{\@biblabel}[1]{\quad#1.}
\makeatother

\usepackage{lastpage,fancyhdr,graphicx}
\usepackage{epstopdf}
\fancyheadoffset[L]{2.25in}
\fancyfootoffset[L]{2.25in}
\begin{document}
\vspace*{0.2in}

\begin{flushleft}
{\Large
\textbf\newline{Artificial intelligence for  topic modelling in Hindu philosophy: mapping  themes between the Upanishads and the Bhagavad Gita} 
}
\newline
\\
Rohitash Chandra \textsuperscript{1 \ddag },
Mukul Ranjan \textsuperscript{2 \ddag },

\bigskip
\textbf{1} Transitional Artificial Intelligence Research Group,  School of Mathematics and Statistics, UNSW, Sydney, Australia
\\ 
\textbf{2} Department of EEE, Indian Institute of Technology Guwahati, Assam, India
\\

\bigskip

%
%

\ddag These authors also contributed equally to this work.



* rohitash.chandra@unsw.edu.au

\end{flushleft}
\section*{Abstract}

A distinct feature of Hindu religious and philosophical text is that they come from a library of texts rather than single source. The Upanishads is known as one of the oldest philosophical texts in the world that forms the foundation of Hindu philosophy.  The Bhagavad Gita is core text of Hindu philosophy and is known as a text that summarises the key philosophies of the Upanishads with major focus on the philosophy of karma. These texts have been translated into many   languages and there exists studies about themes and topics that are prominent; however, there is not much study of topic modelling using language models which are powered by deep learning. In this paper, we use advanced language produces such as BERT to provide topic modelling of the key texts of the Upanishads and the Bhagavad Gita. We analyse the distinct and overlapping topics amongst the texts and visualise the link of selected texts of the Upanishads with Bhagavad Gita. Our results show a very high similarity between the topics of these two texts with the mean cosine similarity of $73\%$. We find that out of the fourteen topics extracted from the Bhagavad Gita, nine of them have a cosine similarity of more than $70\%$ with the topics of the Upanishads. We also found that topics generated by the BERT-based models show very high coherence as compared to that of conventional models. Our best performing model gives a coherence score of $73\%$ on the Bhagavad Gita and $69\%$ on The Upanishads. The visualization of the low dimensional embeddings of these texts shows very clear overlapping among their topics adding another level of validation to our results.

\section*{Author summary}


\section*{Introduction}

  \textit{Philosophy of religion} \cite{meister2009introducing,reese1996dictionary} is a field of study that covers key themes and ideas in religions  and  culture that relate to philosophical topics such as   ethics and metaphysics.   Hindu philosophy \cite{bernard1999hindu,saksena1939nature,chaudhuri1954concept} consists of schools developed for thousands of years which focus on themes such as  ethics\cite{roy2007just},   consciousness\cite{saksena1939nature}, karma\cite{reichenbach1990law,mulla2006karma}, logic and  ultimate reality\cite{chaudhuri1954concept}. Hindu philosophy  is at times  referred as Indian philosophy \cite{dasgupta1975history,radhakrishnan1929indian}. The philosophy of karma and  reincarnation are central to Hindu philosophy \cite{radhakrishnan1929indian}. The Upanishads form the key texts of Hindu philosophy and  seen as the conclusion   of the Vedas 
\cite{staal2008discovering, witzel2003vedas,radhakrishnan1914vedanta,torwesten1991vedanta,prabhu2013mind}. Hindu philosophy \cite{dasgupta1975history} consists of   six major theistic (Astika) schools include  Vedanta \cite{torwesten1991vedanta}, Samkhya \cite{malinar2017narrating}, Nyāya \cite{chakrabarti1999classical}, Vaisheshika \cite{chatterji2007hindu}, Mīmāmsā\cite{arnold2001intrinsic}, 
and Yoga \cite{maas2013concise}. 
Moreover, Jain \cite{long2011jaina}, Buddhist \cite{coomaraswamy2011hinduism,gellner2001anthropology},  Carvaka \cite{bhattacarya2009studies} and Ājīvika  \cite{barua1926ajivika} philosophy are the major agnostic and atheistic (Nastika) schools of Hindu philosophy. 
There has been a lot of interest of Hindu philosophy,   particularly in the west, with a large list  bibliography of translations of key texts such as the Upanishads  \cite{historyUpanisahadtrans}. Moreover, Hindu and Buddhist philosophy have parallels with development of specific themes in Greek philosophy \cite{marlow1954hinduism}.

The  Upanishads and Bhagavad Gita   are  the foundational texts for Hindu philosophy. These texts have been written much later in verse form in Sanskrit language,  they have been sung and remembered for thousands of years  in the absence of a writing system \cite{staal2008discovering}.  The Bhagavad Gita is is part of the Mahabharata which is known as the one of oldest and largest epics written in verse in the Sanskrit language \cite{rajagopalachari1970mahabharata,hiltebeitel1976ritual,gandhi2010bhagavad}. The Bhagavad Gita  is known as a  concise summary of    Hindu philosophy  \cite{radhakrishnan1929indian} with major attribute which is the philosophy of karma   \cite{
brown1958philosophy,phillips2009yoga,muniapan2013dharma}. The Upanishads is a collection of philosophical texts of ancient India which marks the foundation in the history of philosophy \cite{scharfstein1998comparative}. There are 108 key Upanishads of which some were lost in time.  There are 12 are prominent and have been well studied by western scholars  \cite{historyUpanisahadtrans,cross1998turning}.


Nowadays, natural language processing (NLP) methods, that  focuses in processing and modelling language  \cite{indurkhya2010handbook,manning1999foundations,chowdhury2003natural} are typically implemented via  deep learning. NLP considers   tasks such as topic modelling, language translation, speech recognition, semantic and sentiment analysis \cite{manning1999foundations}. Sentiment analysis  provides an understanding of human emotions  and affective states   \cite{liu2012survey,medhat2014sentiment,hussein2018survey}.  Recurrent neural networks such as long-short term memory (LSTM) network models have been prominently used as language models due to their capability to model temporal sequences  \cite{hochreiter1997long}. LSTM models have been improved for language modelling using attention based mechanisms  \cite{wang2016attention}, and    encoder-decoder LSTM framework  with attention (Transformer)  \cite{vaswani2017attention,wolf2020transformers}.  Bidirectional encoder representations from Transformer (BERT) \cite{devlin2018bert} model  is a state-of-art pre-trained language model that features more than 300 million model  parameters for language modeling tasks. Topic models help us better understand a text corpus by extracting the hidden topics.  Traditional topic model such as linear discriminant analysis (LDA) \cite{blei2003latent} assumes that documents are a mixture of topics and each topic is a mixture of words with certain probability score. Sentence BERT (S-BERT)\cite{reimers2019sentence} improves 
BERT model by reducing computational time to to derive semantically meaningful sentence embedding. Recent topic modelling frameworks  use   S-BERT for   embedding in combination with clustering methods \cite{silveira2021topic,peinelt2020tbert,grootendorst10bertopic,angelov2020top2vec,sia2020tired,thompson2020topic}. \textcolor{black}{BERT-based models have shown promising results for topic modelling \cite{grootendorst2022bertopic,peinelt2020tbert,thompson2020topic,glazkova2021identifying}, which motivates their usage in our study. }

 Religious linguistics refer to the study of religious sentences and utterances {\cite{sep-religious-language}}. Major aim of the religious linguistic research is to create an analysis of various subject matters related to religious sentences which include God,  miracles, redemption, grace, holiness, sinfulness along with several other philosophical interpretations
{\cite{pandharipandelanguage,keane1997religious,downes2018linguistics}}.  Most translations of the Bhagavad Gita and related texts come with interpretations and commentary regarding philosophy and  how the verses relate to issues at present \cite{theodor2016exploring}.   Stein \cite{stein2012multi} presented a study about multi-worded expressions by extracting  local grammars based on semantic classes  in the Spanish translation of the Bhagavad Gita and found it to be promising for understanding religious texts and their literary
complexity. The  role of multi-word expressions (MWE) could be a way to better understand metaphorical and lyrical style of the Bhagavad Gita. Rajendran \cite{rajandran2017matter} presented a study on metaphors  in Bhagavad Gita  using   text analysis based on  conceptual metaphor theory (CMT). The analysis identified the source and target
domains for the metaphors,  and traced the choice of metaphors to
physical and cultural experiences. The metaphors
have been inspired by the human body and ancient India, which resonate with modern times. Rajput et al. \cite{rajput2019statistical} provided
a statistical study  of  the word frequency and length distributions prevalent in the translations of Bhagavad Gita  in Hindi, English and French from the original composition in Sanskrit. The Shannon entropy-based measure estimated the vocabulary richness  with Sanskrit as the highest, and  word-length distributions also indicated Sanskrit having  longest word length. Hence, the results demonstrated the inflectional nature of Sanskrit. Dewi \cite{dewi22metaphors} studied  metaphorical expressions and   the conceptual expression underlying in them by reviewing  690 sentences related to metaphor of life from Bhagavad Gita and  analyzed them   using some conceptual metaphor theory. It was reported that the Bhagavad Gita featured 24 conceptual metaphors among which  \textit{life is an entity}, \textit{life is a journey} and \textit{life is a continuous activity} are the most frequent ones. Bhuwak \cite{bhawuk2008anchoring} examined specific ideas from Bhagavad Gita  such as  cognition, emotion, and behaviour by connecting them with the context of human desire. It was reported  that desires lead to behaviour and achievement or non-achievement of desire lead to positive and negative emotions which can be managed in a healthy way by self-reflection, contemplation and the practice of \textit{karmayoga} (selfless action). In our earlier work, BERT-based language model framework was used for sentiment and semantic analysis as a means to compare three different Bhagavad Gita translations where it was found that although the style and vocabulary differ vastly, the semantic and sentiment analysis shows similarity of majority of the verses \cite{ChandraVenkatesh2022}
 
  Although the Bhagavad Gita and Upanishads  have been translated into a number of  languages and   studies about their central themes and topics have been prominent, there is not much work in utilising latest advancements from artificial intelligence, such as  topic modelling using language models -- powered by deep learning. In this paper, we use advanced language models  such as BERT in a framework to provide topic modelling of the key texts of the Upanishads and the Bhagavad Gita. We analysis the distinct and overlapping topics amongst the texts and visualise the link of selected texts of the Upanishads with Bhagavad Gita. Our  major goal is to map the topics in the Bhagavad Gita with the Upanishads; since it is known that the Bhagavad Gita summarizes the key messages in the Upanishads and there are studies about the parallel themes in both texts \cite{haas1922recurrent}. We also provide a comparison of the proposed framework with  LDA which have been prominent for topic modelling.

The rest of the paper is organised as follows.  In Section 2, we provide further details about background  behind  the Bhagavad Gita and Upanishads. Section 3 presents the methodology that highlights model development for topic modelling.  Section 4 presents the results  and Section 5 provides a discussion and future work.

\section{Background}
\subsection{BERT language model}

 BERT  is an attention-based  Transformer model  \cite{vaswani2017attention}  for learning contextualized language representation where  the vector representation of the every input token is dependent on the context of its occurrence in a sentence. The Transformer model \cite{vaswani2017attention} has been  developed by using long short-term memory (LSTM) recurrent neural networks  \cite{hochreiter1997long,greff2016lstm} with an an encoder-decoder architecture \cite{malhotra2016lstm}. Transformer models implement  the mechanism of attention by  weighting the significance of each part of the input data which has been \textcolor{black}{then} prominent for language modelling tasks\cite{vaswani2017attention,beltagy2020longformer}.
 
 BERT is first trained to understand the language (called pre-training phase) and the context after that it is fine-tuned to learn the specific task such as neural machine translation (NMT) \cite{devlin2018bert,imamura2019recycling,yang2020towards,Zhu2020Incorporating,clinchant2019use,shavarani2021better}, question answering \cite{esteva2021covid,khazaeli2021free,geva2021did,ozyurt2020bio,lamm2021qed,kwiatkowski2019natural},and sentiment analysis \cite{hoang2019aspect,li2019exploiting,wu2020context,yang2020cm,du2020adversarial}.  The pre-training phase of BERT involve two different NLP tasks such as masked language modelling (MLM)\cite{devlin2018bert,liu2019roberta,lan2019albert} and next sentence prediction (NSP) \cite{devlin2018bert}. MLM and NSP are semi-supervised learning tasks. In MLM, 15\% words in each input sequence is randomly replaced with a \textit{mask} token and the model is trained to predict these randomly masked input sequences based on the context provided by the neighbouring non-masked words. In NSP, the BERT model learns to predict if two sentences are adjacent to each other. In this way a BERT model is trained simultaneously to minimize the combined loss function, and hence learn the contextualized word embedding. In the fine tuning-phase one or more fully connected layers are added on the top of final BERT layer based upon the applications.  Since BERT is pre-trained, it can be more easily trained further with   datasets for specific applications. In our earlier works, BERT-based framework has been used for   sentiment analysis of COVID-19 related tweets during the rise of novel cases in India \cite{chandra2021covid}. Similar framework using BERT was used for modelling US 2020 presidential elections with sentiment analysis from Tweets in order to predict the state-wise winners \cite{chandra2021biden}.


 Based upon the number of transformer blocks BERT\cite{devlin2018bert} is available with two variants: 1.) $BERT_{BASE}$ consists of 12 transformer blocks stacked on top of each other with a hidden dimension embedding of 768 and 12 Attention heads, on the other hand 2.) $BERT_{LARGE}$ consists of of  24 transformer blocks with a hidden dimension embedding of 1024 and 16 attention heads. $BERT_{BASE}$ has a total of 110 Million parameters while $BERT_{LARGE}$ has a total of 340M parameters.
 BERT takes into account the context for each occurrence of a given word, in comparison to context-free models such as word vectors (word2vec) \cite{mikolov2013efficient} and global vector (GloVe) \cite{pennington2014glove} generate a single word embedding representation for each word in the vocabulary.

\subsection{Document embedding models}

The \textit{universal-sentence-encoder}\cite{cer2018universal} is a sentence embedding model that encodes sentences into high-dimensional embedding vectors that can be used for various natural language processing tasks. The model takes a variable length English text as an input and gives 512-dimensional output vector.The model is trained with deep averaging networks (DANs) \cite{iyyer2015deep} encoder, which simply takes the  average of  the input embeddings for words and bi-grams and then pass them through one or more deep neural networks to get the sentence embeddings. \textit{Sentence-BERT}(S-BERT)\cite{reimers2019sentence} extends the BERT model and Siamese and  triplet network\cite{schroff2015facenet} to generate the sentence embeddings. 
S-BERT uses BERT embeddings with a pooling layer to get the  sentence-embedding ($u$ and $v$) of two sentences. S-BERT has been fine tuned with objective functions such as triplet loss function  and cosine similarity between $u$ and $v$.

\subsection{Clustering techniques}

Clustering is a type of unsupervised machine learning that  groups unlabelled data based on a given similarity measure for a given dataset ${x^{(1)}, ... , x^{(n)}}$,  where $x^{(i)} \in \mathbf{R}^d$ is a d-dimensional data point from the dataset. The goal of clustering  is to assign each data-point a label or a cluster identify. A large number clustering algorithms exits in literature and we used two of them explained below for this work. Xu et al. \cite{xu2015comprehensive} presented an exhaustive list of different groups of clustering algorithms that includes: 1.) centroid based algorithms \textcolor{black}{such as} k-means clustering \cite{steinley2006k}, regards the centroid of data point as the centroid of the corresponding clusters; 2.) hierarchical based algorithms such as agglomerative clustering \cite{gowda1978agglomerative} which creates a hierarchical relationship among the data points in order to cluster them; 3.) density based algorithms that connects an area with high density into clusters \cite{ester1996density}; 4.) distribution based clustering such as Gaussian mixture model \cite{reynolds2009gaussian} that assumes that data generated from same distribution belongs to the same clusters. 


 \textit{K-means clustering}\cite{lloyd1982least} clusters n-data points into k-clusters, where each data point belongs to the cluster with the nearest mean. The k-means algorithm can be explained in the three steps. First step involves initialization of k-centroid corresponding to each clusters. In the second step a point is assigned to a the closest cluster centroid. In the third step, centroid for each cluster is recalculated based on new assigned data points and step 2 and 3 is repeated till convergence.

\textit{Hierarchical density based spacial clustering of application with noise (HDBSCAN}) \cite{campello2013density, mcinnes2017accelerated} is a density-based hierarchical clustering algorithm that defines clusters as highly dense regions separated by sparse regions. The goal of the algorithm is to find  high probability density regions which are our clusters. It starts with estimating the probability density of the data by using the distance of the $k^{th}$ nearest neighbors, defined as the core distance $core_k(x)$. If a region is dense, then the distance of $k^{th}$ nearest neighbor will be less since more data point will fit in the region of small radius. Similarly, for the sparse region,a larger radius would be used. A distance metric called \textit{mutual-reachability-distance} between two points $a$ and $b$ is defined in order to formalize this idea of density and is given by  Equation \ref{eq:mrd}.

\begin{ceqn}
\begin{align}
   d_{\text{mreach-}k}(a, b) = max\{\text{core}_k(a), \text{core}_k(b), d(a, b)\}
   \label{eq:mrd}
\end{align}
\end{ceqn}

where $d(a, b)$ gives the   euclidean distance   between point $a$ and $b$. This mutual reachability distance is used to find the dense areas of the data but since the dense areas are relative and different clusters (dense areas) can have different densities, the entire data points can be modelled as a weighted graph with weight $d_{mreach-k}(a, b)$ of edge between nodes $a$ and b.

\subsection{Dimentionality reduction techniques}
\textit{Uniform manifold approximation and projection(UMAP)}\cite{mcinnes2018umap} for dimension reduction is a non-linear dimensionality reduction technique which is constructed from the theoretical framework based on Riemannian geometry and algebraic topology. The detailed theoretical explanation of the algorithm is out of scope of this paper and can be seen in the paper of McInnes et el.\cite{mcinnes2018umap}. UMAP can be used in a way similar to t-distributed stochastic neighbor embedding (t-SNE) \cite{van2008visualizing}  and principal component analysis (PCA)\cite{wold1987principal} for dimensionality reduction and to visualize  high dimensional data.

Latent dirichlet allocation (LDA)\cite{blei2003latent} is a generative probabilistic model for the topic modelling of the corpus based on word frequency. The basic idea behind the model is that, each document is generated by a statistical generative process and hence each document can be modelled as a random mixture of latent topics and each topic is mixture of words characterised its distribution. 
A \textit{word} denoted by $w$ and indexed from $1$ to the vocabulary size $V$ and a \textit{document} is given by $\mathbf{w} = \{w_1, w_2, ..., w_N\}$, where $w_i$ is the $i^{th}$ word in the sequence \cite{blei2003latent}. The generative process involved in the algorithm can be summarized as 1.) fix the number of topic and hence the dimensionality of the Dirichlet distribution and that of the topic variable $z$ and sample $\theta$(per-document topic proportion) from a Dirichlet prior $Dir(\alpha)$ 2.) sample a topic $z_n$ from a multinomial distribution $p(\theta; \alpha)$ and then 3.) sample a word $w_n$ from multinomial probability distribution conditioned on $z_n$, $p(w_n|z_n, \beta)$. Overall probability of document $\mathbf{w}$ containing $N$ words can be given by Equation \ref{eq:pw}.

\begin{ceqn}
\begin{equation}
  p(\mathbf{w})=\int_{\theta}\left(\prod_{n=1}^{N} \sum_{z_{n}=1}^{k} p\left(w_{n} \mid z_{n} ; \beta\right) p\left(z_{n} \mid \theta\right)\right) p(\theta ; \alpha) d \theta 
  \label{eq:pw}
\end{equation}
\end{ceqn}

Given a corpus of $M$ documents $D = \{w_1, ..., w_M\}$, the EM algorithm can be used to learn the parameters of an LDA model by maximizing a variational bound on $p(D)$, as seen in Equation \ref{eq:lpd}.
\begin{ceqn}
\begin{equation}
\log p(\mathcal{D}) \geq \sum_{m=1}^{M} \mathrm{E}_{q_{m}}[\log p(\theta, \mathbf{z}, \mathbf{w})]-\mathrm{E}_{q_{m}}\left[\log q_{m}(\theta, \mathbf{z})\right]
\label{eq:lpd}
\end{equation}
\end{ceqn}

\textcolor{black}{LDA has been used for several language modeling tasks that include the study of the relationship between two corpora using topic modeling\cite{ding2019overlaplda} which is also the focus of our study. }

\section{Methodology} 
 
 \subsection{Datasets}

We evaluated a number of prominent translations of the  Bhagavad Gita and the Upanishads. In order to maintain the originality of the themes and ideas of these two classical Indian texts, we   used the older and the most popular translations for this work. We chose Eknath Eashwaren's translation since he directly translated from Sanskrit to English and translated both texts \cite{easwaran2007bhagavad,easwaran1987upanishads}, hence it would be not be creating a translation bias for topic modelling and comparison of the topics between the texts. Eknath Easwaran ( 1910 – 1999) was a professor of English literature in India and later moved to the United States where he translated these texts.  In addition, we chose the translation by Shri Purohit Swami and  William Butler  Yeats \cite{swami2012ten} for further comparison.   W. B Yeats (1865 –   1939) was Irish poet,  dramatist, prose writer and known as  one of the foremost figures of 20th-century literature.  Shri Purohit Swami (1882 – 1941) was a Hindu teacher from Maharashtra, India. The translation of the Upanishads by them is special since it has been \textcolor{black}{done} jointly by prominent Indian and Irish scholars and captures   Eastern and Western viewpoints. Table \ref{tab:texts} provides further details of the texts.  Note that Shri Purohit Swami also translated the Bhagavad Gita \cite{swami_purohit} which can be used in future analysis, and not used in this work.

\begin{table*}
\small
\centering
\begin{tabular}{l  l l }
\hline
Texts & Translator & Year\\ \hline
\hline
The Bhagavad Gita\cite{easwaran2007bhagavad} &Eknath Easwaran &  1985\\ 
\hline
The Upnishads\cite{easwaran1987upanishads} & Eknath Easwaran & 1987\\
The Ten Principal Upanishads\cite{swami2012ten} & Shri Purohit Swami \& W.B. Yeats & 1938\\
108 Upanishads\cite{108upanishads} & The International Gita Publication & --\\ 
\hline
\end{tabular}
\caption{Details of the texts used for topic modelling.}
\label{tab:texts}
\end{table*}


The Bhagavad Gita consist of 18 chapters which features a series of questions and answers between Lord Krishna and Arjuna that range with a range of topics that includes the philosophy of Karma. The Mahabharata war lasted for 18 days \cite{thapar2009war}; hence, the organisation of the Gita is symbolic.  

 

\textit{The Upanishads}\cite{easwaran1987upanishads} translated by Eknath Eashwaren  provides  a commentary and translation  of the 11 major and 4 minor Upanishads. The \textit{108 Upanishads}\cite{108upanishads}   is a collection of the translation and commentary of all 108 Upanishads in a single book compiled by the \textit{Gita Society}. The translation and commentary is done by a group of  spiritual teachers who have tried to recover the Upanishads which have believed to be lost earlier; however, there are not much details about how they have recovered them  \cite{108upanishads}.  The Chandogya Upanishad has highest number of words followed by the Katha Upanishad and the Brihadaranyaka Upanishad. \textit{The Ten Principal Upanishads}\cite{swami2012ten} consists of the translation of the 10 major Upanishads. This text does not have a separate explanation for each Upanishad unlike the Upanishads by Eknath Easwaran. The Brihadaranyaka Upanishad consists of the highest number of words followed by the Chandogya Upanishad and Katha Upanishads.  The Chandogya Upanishad is one of the largest Upanishads consisting of 8 chapters which can be divided into 3 natural groups according to the philosophical ideas \cite{witz1998supreme}. The first group (Chapter 1 and Chapter 2) deals with the structure and different aspects of the languages and its expression, particularly with the syllable "Om" that is used to describe Brahman and beyond. The second group (Chapter 3-5) consists of the ideas of universe, life, mind and spirituality. The third group (Chapter 6-8) deals with the more metaphysical questions such as nature of reality and Self \cite{witz1998supreme}. Since first five chapters are   intermixed with rituals, Shri Purohit Swami  omitted them from  in his translation \cite{swami2012ten} along with some passages from the Brihadaranyaka Upanishad. Other authors also state that some of the passages of the Brihadaranyaka Upanishad has been omitted due to the repetitions \cite{swami2012ten}. Brihadaranyaka Upanishad, consisting of 6 chapters discusses about different philosophical ideas including one of the earliest formulation of the Karma doctrine (Verse 4.4.5), ethical ideas such as self-restraint (Damah), charity (Danam) and compassion (Daya) and also other metaphysica topics related to philosophy of Advaita Vedanta. Eknath Easwaran \cite{easwaran1987upanishads} translated this chapter as the \textit{Forest of Wisdom} which starts with the one of Vedic theories of the creation of the Universe and then the dialogue between a great sage, Yajnavalkya, and his wife Maitreyi which is  a deep spiritual discussion about death, possession, self, Brahman (God) and the Atman (Self). It contains one of the earliest psychological theories relating the human body, mind, ego and the Self. The Katha Upanishad is one of the legendary story of a small boy Nachiketa who met Yama (the god of Death) asks with him different questions about the nature of life, death, man, knowledge, Atman  and Moksha (liberation). The Katha Upanishad consists of 2 chapters each consisting of 3 sections.

\subsection{Framework}


Our  major goal is to map the topics in the Bhagavad Gita with Upanishads. We begin by selecting 12 prominent Upanishads \textit{(Isha, Katha, Kena, Prashna, Munda, Mandukya, Taittiri, Aitareya, Chandogya,  Brihadaranyaka, Brahma, Svetasvatara)} from the text translated by Eknath Easwaran\cite{easwaran1987upanishads}. The major reason that we selected both by the same author for this task is to eliminate any bias in translation for topic modelling. However we also considered other translations as mentioned in the table \ref{tab:texts} and found that these bias does affect the similarity matrix. For example when we compared the similarity between the Upanishads by Eknath Easwaran and the Bhagavad Gita by the same authors, average similarity score is 3\% better than that of the Bhagavad Gita by Eknath Easwaran and the Upanishads by Shri Purohit Swami. Finally, we also presents the visualization of the topics space of 108 Upanishads and its different part divided based on the original Vedas these Upanishads are originated from..

\begin{figure*}

\begin{adjustwidth}{-2.25in}{0in}
\includegraphics[width=.99\linewidth]{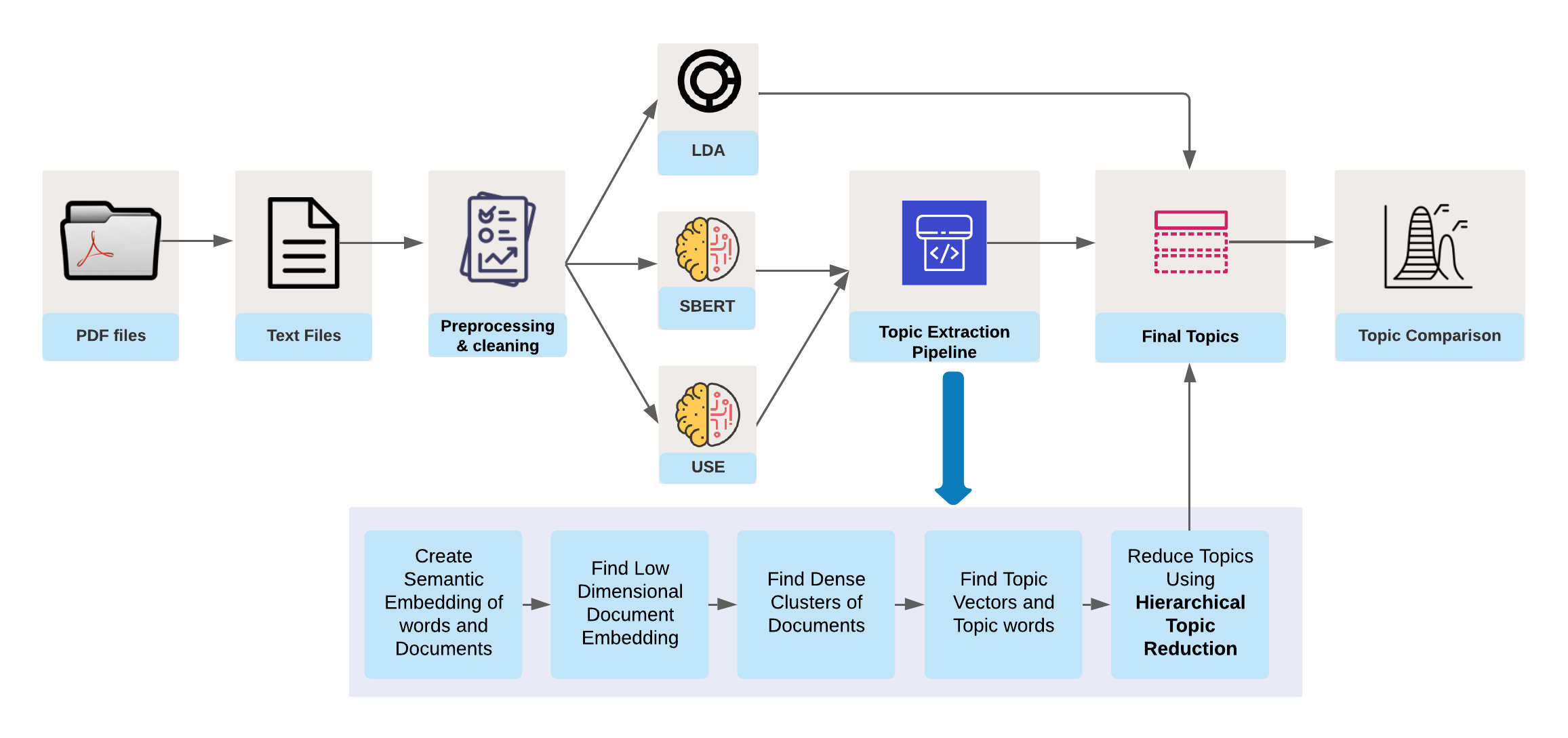}
\caption{Framework for semantic similarity and Topic Modelling }
\label{fig:framework}

\end{adjustwidth}
\end{figure*}

Next, we present a framework that employs different machine learning methods for topic modelling. Figure \ref{fig:framework} presents the complete framework for the analysis and topic modelling  of the respective  texts given in Table \ref{tab:texts}.  In Figure \ref{fig:framework}, the first stage consists of conversion of PDF files and text pre-processing as discussed in the previous section. In the second stage, we use  two  different sentence embedding models  1.) universal sentence encoder (USE) and 2.)  Sentence-BERT(SBERT) for  generating the word and documents embedding which is later passed thorough the topic extraction pipeline  to generate the topic vector and  finally we compared our results with the classical topic modelling algorithm LDA\cite{blei2003latent} across different corpus.
Our framework to generate topics is similar to Top2Vec \cite{angelov2020top2vec}, however we also used other clustering algorithms like K-Means. First, USE and SBERT are  used to generate the joint semantic embeddings of documents and words. Since these embeddings are generally in higher dimension which is very sparse, we need to reduce the dimension of the embeddings to get the dense areas. We use dimensionality reduction techniques like UMAP and PCA for reducing the high dimensional embedding vectors generated by the S-BERT and the USE. Next, we find dense clusters of topics in the document vectors of the corpus using  clustering algorithms like HDBSCAN and K-Means. These clusters are represented by the centroid of document vectors in the original dimension, which is called as topic vectors\cite{angelov2020top2vec}. Next, we find top N(N = 50 in our case) nearest words to the topic vectors which represent our final topic. Topic vectors also allows us to group the similar topics and hence reduce the number of topics using  Hierarchical Topic Reduction\cite{angelov2020top2vec}.

Most of the topic modelling research \cite{doan2021benchmarking,angelov2020top2vec,lisena2020tomodapi} involves the bench-marking model results on pre-existing datasets such as the \textit{20 News Groups dataset}\cite{lang200820}, the \textit{Yahoo Answers dataset} \cite{yin2019benchmarking, zhang2015character}, \textit{Web Snippets dataset}\cite{ueda2003parametric}, \textit{W2E datasets}\cite{hoang2018w2e}. These datasets have been prepared to be used for the algorithms bench-marking tasks and consists of the fixed number of documents and words. The \textit{20 News Groups Datasets} for example consists of 15,465 documents and 4,159 words \cite{doan2021benchmarking}. Tweets have also been used for topic modelling tasks\cite{jonsson2015evaluation, yan2013biterm, sridhar2015unsupervised}. Jonsson et al. \cite{jonsson2015evaluation} for example, collected tweets from Twitter to prepare a datasets of 129,530 tweets and used  LDA \cite{blei2003latent}, \textit{Biterm-Topic-Model}(BTM)\cite{yan2013biterm} and a variation of LDA algorithms for topic modelling to compare their performance. In case of Twitter based topic modelling datasets, a tweet is considered as ${Document}$, though Jonsson et al. \cite{jonsson2015evaluation} aggregate documents to form \textit{pseudo-documents} and found that it solves the poor performance of LDA on shorter documents.
Murakami et al. \cite{murakami2017corpus} used research papers published in the journal \textit{Global Environmental Change (GEC)} from the first volume (1990/1991) to Volume 20 (2010) as the corpus for the topic modelling. They divided the a paper into several paragraph blocks and modelled them as a documents of the corpus. 

Our dataset can be seen as similar to Murakani et al.\cite{murakami2017corpus}. The Bhagavad Gita and Upanishads are written in verse form and to maintain the originality of the texts, most of the translations also preserve the numbering of the verses. Other than the verses, the translations also contain commentary by the translator of the texts. While creating the datasets, we first created documents based on the verse number in the texts, i.e a verse is considered as a \textit{document of the corpus}, where the numbering are clearly mentioned. In other cases when verse numbers are not mentioned clearly, we considered one paragraph as one documents. In case of the commentary, we split the commentary into smaller parts to make them a \textit{document} as done by Murakami et al.\cite{murakami2017corpus}.
The statistics in terms of number of documents, number of  words  (\# words), average number of words (avg \# words), and number of verses (\# verses) of different corpus (text files) and their details can be found in Table \ref{tab:corpus_stat}.

 \subsection{Text data extraction and processing}
 
In order to process the files given in printable document format (PDF), we converted them into text files. Most of the PDF files were generated from the scanned images of the printed texts, hence we used optical character recognition (OCR) based open-source library\footnote{\url{https://github.com/writecrow/ocr2text}}. This conversion from PDF to text file gave us a raw dataset consisting of all the texts shown in Table \ref{tab:texts}. Next, pre-processing done on the entire datasets, which consists of the following steps.
\begin{enumerate} 
    \item Removing unicode characters generated in the text files due to noise in the PDF files;
    \item Normalizing(assigning uniform verses from each text) verse numbering in the Upanishads and the Bhagavad Gita;
    \item Replacing the archaic English words such as "thy" and "thou" with modern English words like your and you;
    \item Removing the punctuation, extra spaces, and lower-casing;
    \item Removing repetitive and redundant sentences such as "End of the Commentary".
\end{enumerate}

 Examples of selected text from the original document  along with the processed text is shown in Table \ref{tab:cleaning}. The original text and processed text has been given in the Github repository \footnote{\url{https://github.com/sydney-machine-learning/topicmodelling_vedictexts/}}. In topic modelling literature, \textit{word} is the basic unit of data which is defined to be an items from vocabulary indexed by $\{1, ..., V\}$, where $V$ is the vocabulary size. A \textit{Document} is a collection of $N$ words represented by $\mathbf{w} = \{w_1, w_2, ..., w_N\}$, where $w_i$ is the $i^{th}$ word in the sequence. The corpus is considered as a collection of $M$ documents denoted by $D = \{\mathbf{w}_1, \mathbf{w}_2, ..., \mathbf{w}_M\}$\cite{blei2003latent}.

\begin{table*}
\begin{adjustwidth}{-2.25in}{0in}
\small 
\begin{tabular}{ll}
\toprule
\hline
\hline
Original Documents & Transformed Documents  \\ \hline
\hline
\begin{tabular}[c]{@{}l@{}}II-5(a). What winds up empirical life is (its) appearance \\ as unreal. \end{tabular} & \begin{tabular}[c]{@{}l@{}}what winds up empirical life is its appearance as unreal. \end{tabular} \\

\hline 
\begin{tabular}[c]{@{}l@{}}"What discipline is required to know, \textbackslash u2018this is a pot,\\ except the adequacy of the means of right \textbackslash u2019 \textbackslash n \\knowledge ?" \end{tabular} & \begin{tabular}[c]{@{}l@{}} "what discipline is required to know this is a pot except\\ the adequacy of the means of right knowledge."\end{tabular} \\
\hline 

\hline 
\begin{tabular}[c]{@{}l@{}}Lord, have we not prophesied in thy name? and in thy \\name have cast out \textbackslash n devils? and in thy name done \\many wonderful works? \end{tabular} & \begin{tabular}[c]{@{}l@{}} Lord have we not prophesied in your name and in\\ your name have cast out devils and in your name\\ done many wonderful works. \end{tabular} \\
\hline 

\end{tabular} 
 \caption{Processed text after removing special characters and transforming archaic words into modern English}
\label{tab:cleaning}
\end{adjustwidth} 
\end{table*}

\subsection{Technical details}


In our framework, S-BERT and USE are used for the task of generating sentence embeddings. We used pre-trained S-BERT\footnote{\url{https://huggingface.co/sentence-transformers/}}, which has been trained on a large multilingual corpus. The model uses DistilBERT\cite{sanh2019distilbert} as the base transformer model, then its output is pooled using an average pooling layer and a fully connected (dense) layer is used finally to give a 512 dimensional output. We used different combination of dimensionality reduction techniques and clustering algorithms with the pre-trained semantic embeddings to get the final topics for each corpus.

The embedding dimension is reduced to the 5-dimension using the selected dimensionality reduction techniques i.e UMAP and  PCA.  UMAP uses two important parameters, \textit{n\_neighbors} and \textit{min\_dist} in order to control the local and global structure of the final projection. We fine-tuned these parameters to optimize the topic-coherence metric and use the final UMAP model with the default \textit{min\_dist} value of $0.1$, \textit{n\_neighbors} value of $10$ and the \textit{n\_components} value of $5$, which is the final dimension of the embeddings. We set the \textit{random-state} to $42$ and use \textit{cosine-similarity} as the distance metric.  

After getting the embedding of the documents in the reduced dimensions we used two different clustering algorithms - HDBSCAN\cite{campello2013density, mcinnes2017accelerated} and K-Means\cite{lloyd1982least} algorithms to cluster the documents where each clusters represent a topic. 
We fine-tuned different parameters of HDBSCAN to get the optimal value of topic coherence metric(Topic coherence is discussed with great details in  next section), which represents how good our generated topics are.\textcolor{black}{We chose the number of topics obtained at the optimal value of topic coherence metric as the optimal number of topics and used the same number as the value of $K$ for K-Means Clustering algorithms}.
The  \textit{min\_cluster\_size}   defines the smallest grouping size to be considered as cluster, we set it to $10$. Finally, in the remaining two parameters,  we use  $metric=euclidean$ and $min\_samples = 5$. The k-means algorithm is trained for the  300 iterations, with the num\_clusters parameters same as the number of labels found using HDBSCAN. 

 \section{Results}

\subsection{Data Analysis} 
We begin by reporting key features of the selected texts (datasets)  as shown in Table \ref{tab:corpus_stat}. \textit{The Upanishads} by Eknath Easwaran contains 862 documents, 40737 words and 705 verses. Since this text contains explanation by the authors as well so the number of documents is more than the number of verses for this text. \textit{Ten Principal Upanishads} by W.B. Yeats and Shri Purohit Swami Consists of 1267 documents and same number of verses as well. The corpus also consists of 27492 words with an average of $21.70$ words per documents. \textit{The Bhagavad Gita} by Eknath Easwaran consists of 700 verses and the same number of documents along with 20299 words with an average of 21.70 words per documents.

\begin{table*}
\small
\centering
\begin{tabular}{||c |c |c |c |c ||}
\hline
Corpus & \# Documents & \# Words & Avg \# words & \# Verses\\ \hline
\hline
The Upanishads(Eknath Easwaran) & 862 & 40737 & 47.26 & 708\\
\hline
The Bhagavad Gita(Eknath Easwaran) & 700 & 20299 & 27.50 & 700\\
\hline
Ten principal Upanishads & 1267 & 27492 & 21.70 & 1267\\



\hline
108 Upanishads & 6191 & 405559 & 65.50 & 6191\\

\hline
\end{tabular}
\caption{Dataset Statistics}
\label{tab:corpus_stat}
\end{table*}


\begin{figure*}

\begin{adjustwidth}{-2.25in}{0in}
\includegraphics[width=.59\linewidth]{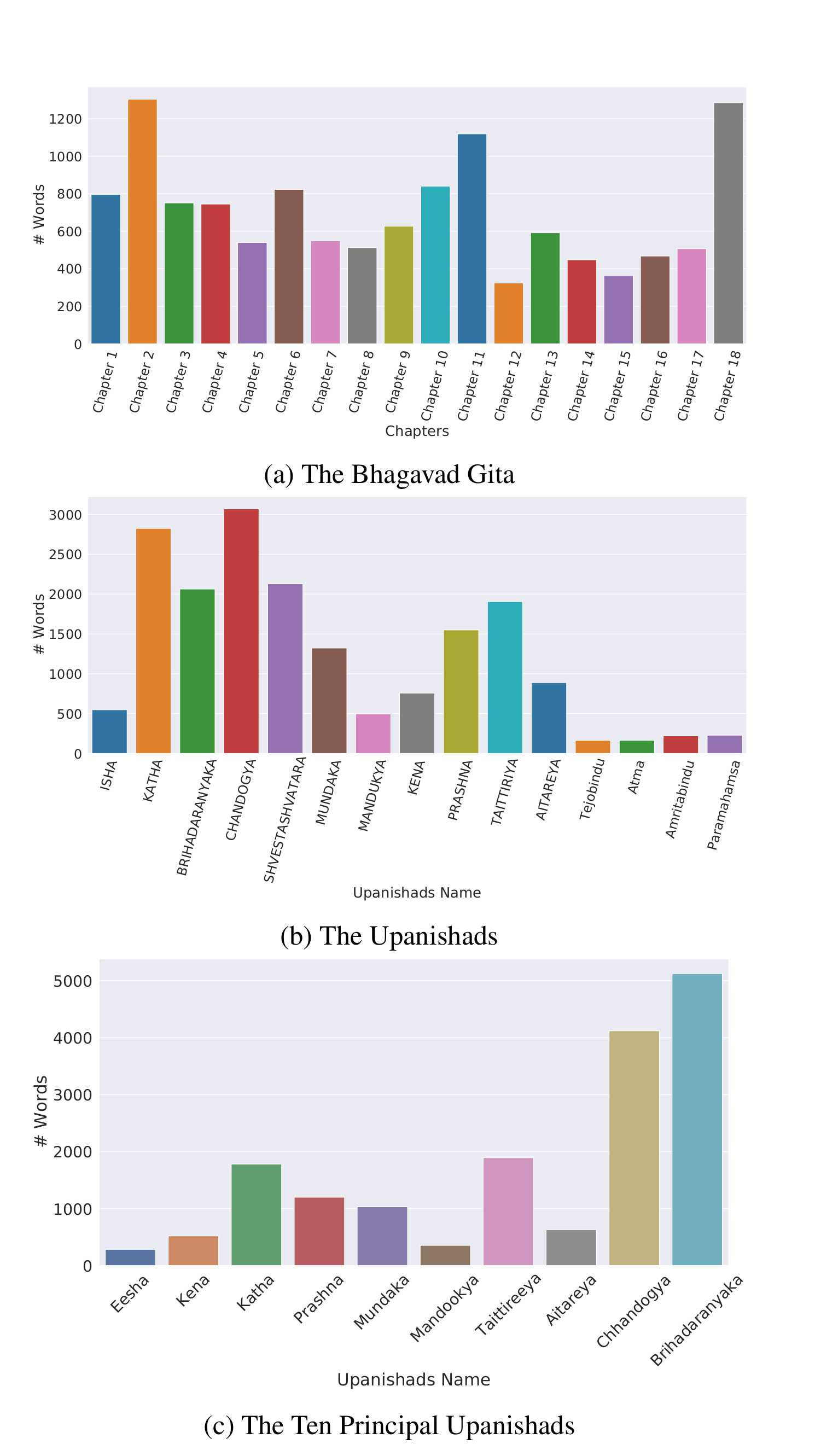}
\caption{Chapter Wise Word count for different texts in the dataset}
\label{fig:chapter_wise_wc}

\end{adjustwidth}
\end{figure*}

Figure \ref{fig:chapter_wise_wc} shows the chapter-wise word count of the respective corpus. \textit{The Bhagavad Gita} consists of 18 chapter  and we see that Chapter 2 has highest number of words, followed by Chapter 18 and Chapter 11.
This is because these chapters contains relatively more number of verses and explains much deeper topics of Hindu philosophy. Chapter 18 of the Bhagavad Gita contains the highest number of verses(78) followed by the chapter 2 which contain 72 verses and chapter 11 containing 55 verses. Chapter 2 of the Bhagavad Gita discuss about the Samkhya and Yoga School of Hindu  Philosophy\cite{chinmayananda1996holy,yogananda2007god,easwaran2007bhagavad}. It teaches about the cosmic wisdom (Brahm Gyan) and the methods of it attainment along with along with the notion of qualia (Atman/self), duty, action (karma), selfless action (karma yoga), rebirth,  afterlife, and the qualities of self-realized individuals (muni) \cite{yogananda2007god}. Eknath Easwaran\cite{easwaran2007bhagavad} claimed this chapter as an overview for the remaining sixteen chapters of the Bhagavad Gita. Chapter 11 is also called as the "Vishwa Roopa Darshana Yoga" \cite{chinmayananda1996holy} which has been translated as "The Cosmic Vision" by Eknath Easwaran \cite{easwaran2007bhagavad}, and "The Yoga of the Vision of the Universal" Form\cite{chinmayananda1996holy} by Swami Chinmayananda. This chapter talks about the supreme vision of Lord Krishna which made Arjuna experience deep peace and joy of Samadhi (enlightenment) along with the feeling of being terrified at the same time \cite{yogananda2007god,easwaran2007bhagavad}. When terrified, Arjuna asks about the identity of the cosmic vision of God. Lord Krishna replies in verse-32 of Chapter 11 that came into Robert Oppenheimer’s mind when he saw the atomic bomb explode over Trinity in the summer of 1945 \cite{easwaran2007bhagavad,bera2018oppenheimer}.  He mentioned, \textit{"A few people laughed, a few people cried. Most people were silent. I remembered the line from the Hindu scripture, the Bhagavad Gita; Vishnu is trying to persuade the Prince that he should do his duty and, to impress him, takes on his multi-armed form and says, Now I am become Death, the destroyer of worlds."}



\begin{figure*}[hbp!]
    \centering
    \includegraphics[width=\textwidth]{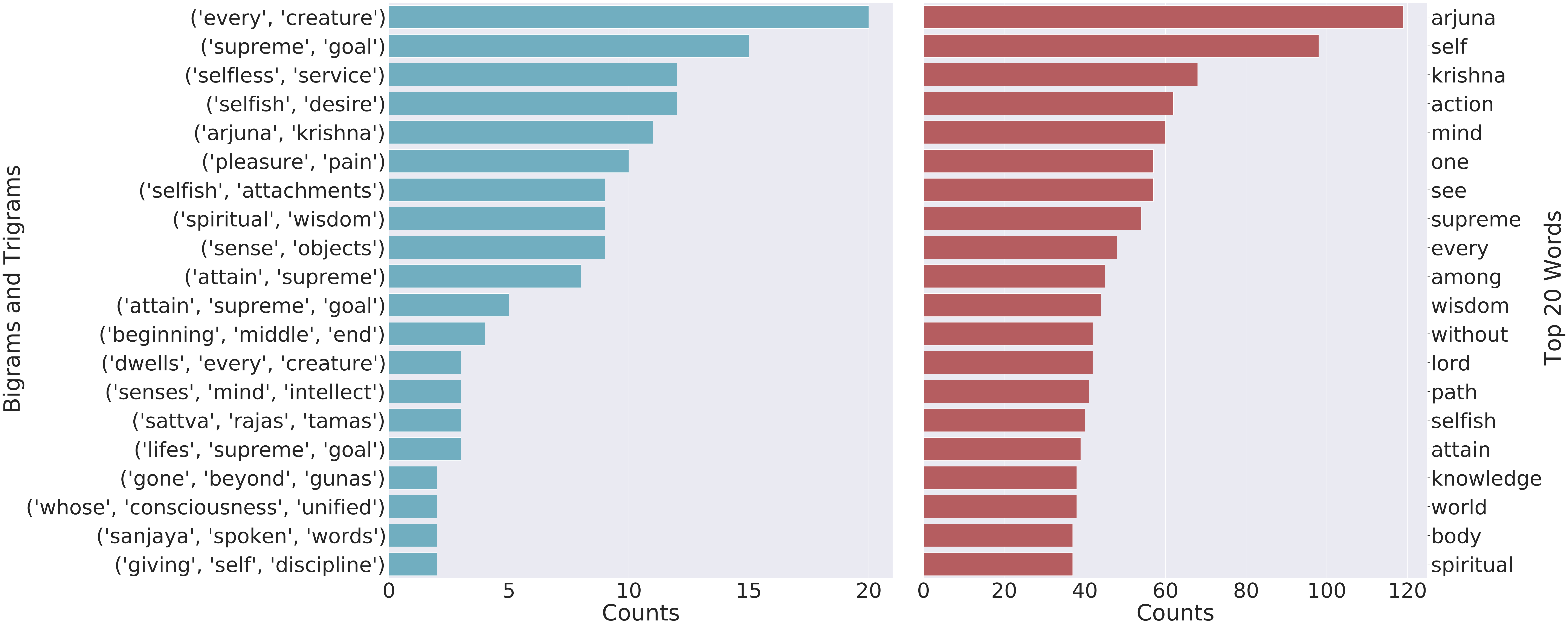}
    \caption{Visualisation of top 20 words, and top 10 bigrams and trigrams for the Bhagavad Gita.}
    \label{fig:eknath_gita_uni_bi_tri}
\end{figure*}

\begin{figure*}[hbp!]
    \centering
    \includegraphics[width=\textwidth]{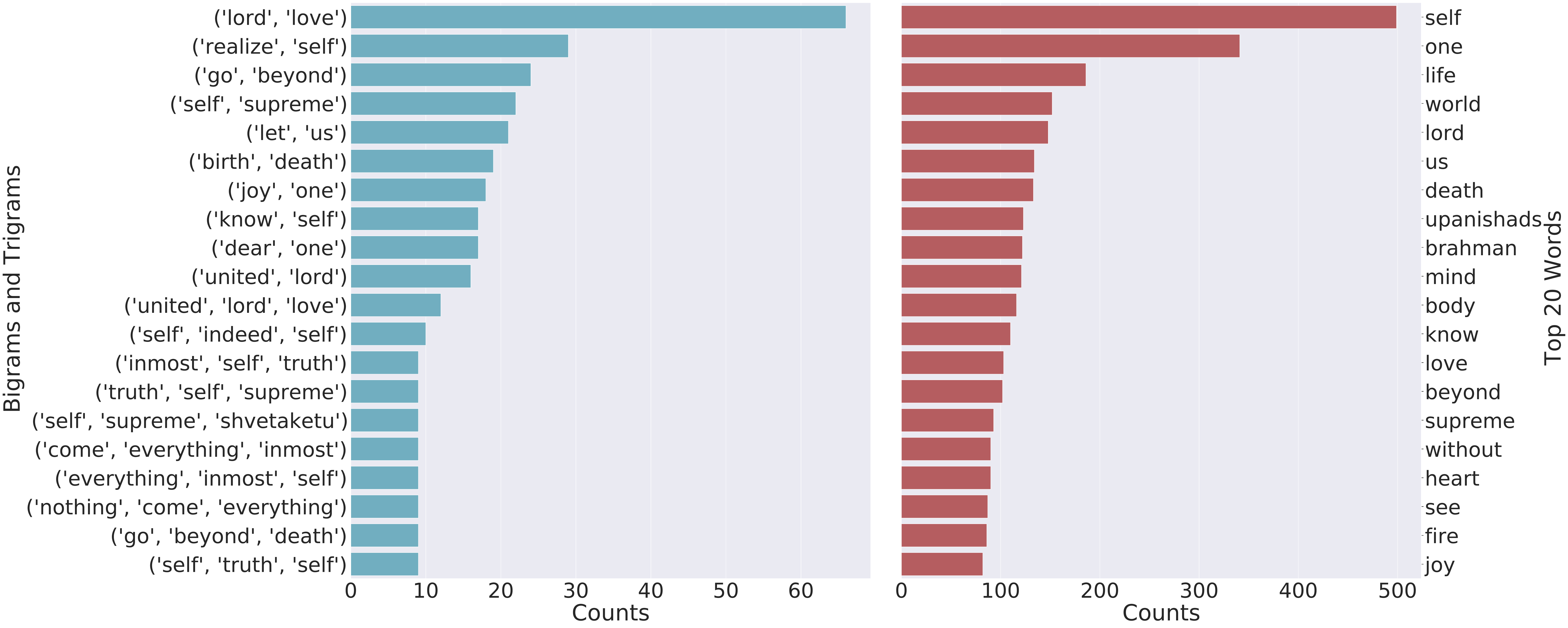}
    \caption{Visualisation of top 20 words, and top 10 bigrams and trigrams for the Bhagavad Gita. Upanishads by Eknath Easwaran.}
    \label{fig:eknath_upanishads_uni_bi_tri}
\end{figure*}

\begin{figure*}[hbp!]
 
\begin{adjustwidth}{-2.25in}{0in}
    \includegraphics[width=\textwidth]{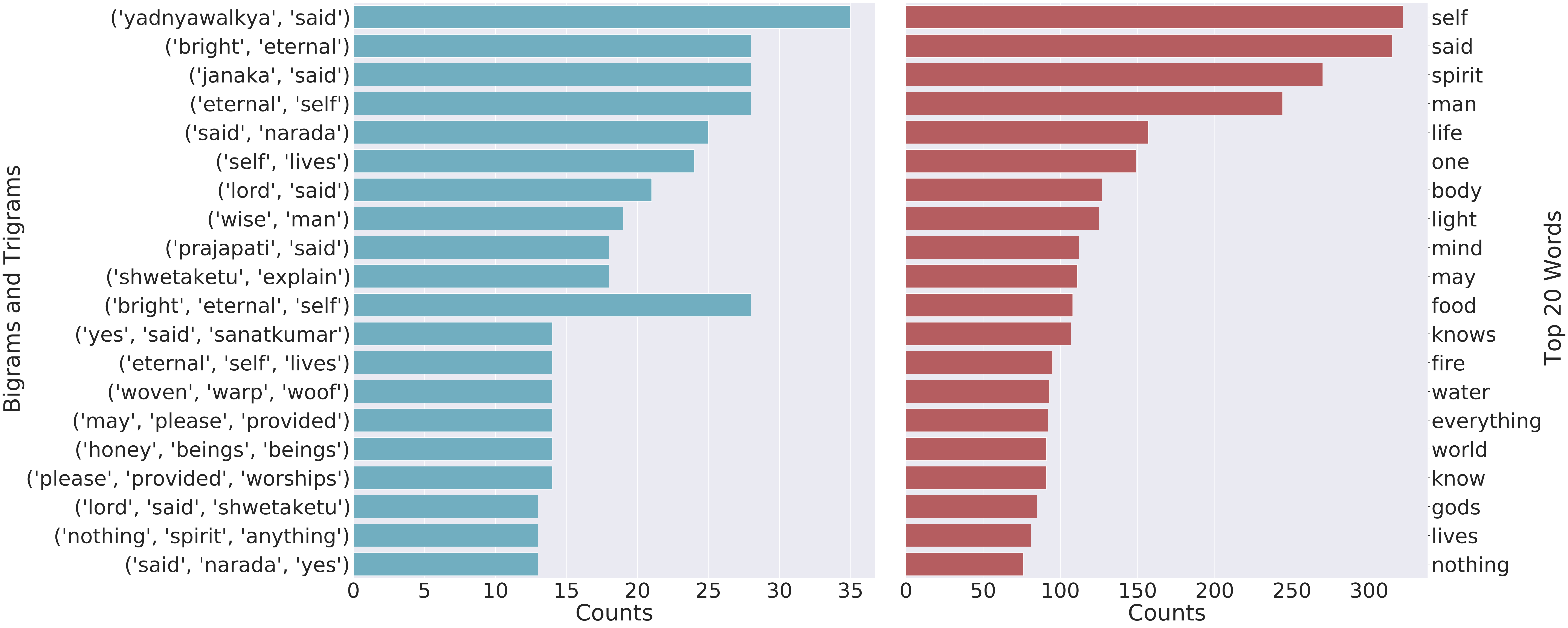}
    \caption{Visualisation of top 20 words, and top 10 bigram and trigrams for the ten principal Upanishads.}
    \label{fig:ten_upanishads_uni_bi_tri}
    
\end{adjustwidth} 
\end{figure*}

The n-gram\cite{sidorov2014syntactic} is typically used to provide basic statistics  of a text using a continuous sequence of words or other elements.  Bi-grams and tri-grams are typical examples of n-grams. Figure \ref{fig:eknath_upanishads_uni_bi_tri} shows the count of top 10 bigrams and trigram along with the top 20 words for the Upanishads by Eknath Easwaran.   In the case of the Upanishads by Eknath Easwaran, (lord, love) is the most frequent bigram which has occurred  more than 60 times followed by (realize, self) and (go, beyond). In the same corpus, when we look at the trigram's bar plot we find that (united, lord, love), (self, indeed, self) and (inmost, self, truth) to be the top 3 trigrams of the corpus. Similarly Figure \ref{fig:ten_upanishads_uni_bi_tri} shows  unigrams, bigrams and the trigrams of the Ten Principal Upanishads.
Although these n-grams just state the frequency of occurrence of the continuous sequence of words, they give a rough idea about the themes and topics discussed in the corpus. This can be seen in Figure \ref{tab:topic-ten-upan-docs} that a lot of topics do contain these words. We can see that 'self' is one of the predominant word of topic 4 and topic 8 of the ten principal Upanishads. In fact when we observe these topics carefully, we find that the entire topic is related to the theme of 'self'. Similarly, we find words 'lords', 'god' and 'sage' to be predominant words in topic 1 and topic 3 of the ten Principal Upanishads.
Figure \ref{fig:eknath_gita_uni_bi_tri} shows the bigrams, trigrams and word count for the Bhagavad Gita. We see that 'arjuna', 'self', 'krishna', 'action' and 'mind' are top 5 words of the Bhagavad Gita. Among the bigrams and trigrams, we see that (every, creature), (supreme, goal) and (selfless, service) are top 3 bigrams while (attain, supreme, goal), (beginning, middle, end) and (dwells, every, creature) are the top 3 trigrams of the text. Since Arjuna and Krishna are the protagonists of the text, it is obvious for them to be among the top words of the text. We see that other than  \textcolor{black}{these}, 'self, action, and mind' are the prominent words  that give us a basic idea about the themes   that can be verified from the topics presented in Figure \ref{tab:topic-gita-docs}. Topic 1 of the Bhagavad Gita shown  in Figure \ref{tab:topic-gita-docs} shows all the names of the Hindu spiritual entities (deities). We find that Krishna and Arjuna are one of the major one among them. This topic also includes other entities and deities such as Jayadratha, Vishnu and Bhishma that have been mentioned by the Lord Krishna  in the text. The words related to the 'self' can be seen in topic 2 of  Figure \ref{tab:topic-gita-docs}; hence, we can conclude that themes related to self are present in Topic 2 identified by our framework. We can also see that topic 13 of the Bhagavad Gita contains the words related to 'action' (karma)  which is also one of the top 5 words of the texts.

In terms of the individual word frequency,  we find that "self" is one of the most occurred word in all the three corpus which is a major theme of Hindu Philosophy. The "self" is the translation from the Sanskrit word "Atman", which refers to the spirit, and more precisely "qualia" as known in the definition pertaining to the \textit{hard problem of consciousness} \cite{stubenberg1998consciousness}. The "Atman" is also often translated as consciousness and there are schools of thought (Advaita Vedanta \cite{indich1995consciousness}) that sees the Atman as Brahman (loosely translated to the concept of God or super-consciousness) \cite{
woodhouse1978consciousness, saksena1939nature}. Often, it is wrongly translated to the term soul which is an Abrahamic religious concept, where humans  only to have the soul and the rest of  life forms do not  \cite{ackerman1962debate}. Atman, on the other hand, is the core entities of all life forms and also of non-life forms in Hindu philosophy. Not only in Upanishads but it has been explained in The Bhagavad Gita as well with a great details. Finally,  "attain supreme goal" is the most occurred trigram of the Bhagavad Gita   which suggests that the Bhagavad Gita talks about attaining supreme goal with a great details along with the other philosophical topics. The Bhagavad Gita is also known as the \textit{Karma Upanishad} or the text that focuses on philosophy of karma (action/work) \cite{mulla2006karma}. The major focus of the Bhagavad Gita is karma philosophy given a conflicting situation and the path to self realisation as the goal of life, and hence, it has also been recognised as a book of leadership and management \cite{sharma1999corporate,nayak2018effective}, and psychology \cite{reddy2012psychotherapy}.

\subsection{Modelling and Predictions}
\subsubsection{Topic Coherence}

Quantitative evaluation of the topic model is one of its major challenge. Initially, topic models were evaluated with held-out-perplexity but it does not correlate with the human evaluation \cite{chang2009reading}. A topic can be said to be coherent if all or most of the words of the topic support each other or are related\cite{syed2017full}. Human evaluation of topic coherence is done in two ways: 1.) rating, where human evaluators rate the topic quality on a three point topic quality score, and 2.) intrusion, where each topic is represented by its top words along with an  intruding word which has very low probability of belonging to the topic since it does not belong in the topics uncovered. It is a behavioral way to judge topic coherence and measured by how well a human evaluator can detect the intruding word \cite{chang2009reading, morstatter2018search}. 
Automated topic coherence metric based on \textit{normalized pointwise mutual information}(NPMI) correlates really well with the human evaluation and interpretation of the topic coherence \cite{newman2010automatic, newman2010evaluating, newman2011improving, morstatter2018search}. R{\"o}der et. al.\cite{roder2015exploring} provided a detailed study on the coherence measure and its correlation with the human topic evaluation data. We use the \textit{topic coherence} NPMI measure(TC-NPMI)\cite{roder2015exploring} as a metric to fine tune and evaluate different models on different corpus. Equation \ref{eqn:npmi} gives the NPMI for a pair of   words $(w_i, w_j)$ from the top N (set to 50) words of a given topic: 

\begin{ceqn}
\begin{equation}
  \operatorname{NPMI}\left(w_{i}, w_{j}\right)=\left(\frac{\log \frac{P\left(w_{i}, w_{j}\right)+\epsilon}{P\left(w_{i}\right) \cdot P\left(w_{j}\right)}}{-\log \left(P\left(w_{i}, w_{j}\right)+\epsilon\right)}\right)
  \label{eqn:npmi}
\end{equation}
\end{ceqn}

where, the joint probability $P\left(w_{i}, w_{j}\right)$, i.e  the probability of the single word $P\left(w_{i}\right)$ is calculated by the Boolean sliding window approach (window length of $s$ set to the default value of 110). A virtual document is created and the count of occurrence of the word ($w_i$) or the word pairs ($w_i, w_j$), and then it is divided by the total number of the virtual documents.

We use TC-NPMI  as the topic-coherence measure to evaluate different topic models and tune different hyper-parameters of different algorithms. Table \ref{tab:metric} shows the value of metric for different model on different datasets. We trained the LDA model  for 200 iterations with other hyper-parameters set to the default value as given in the \textit{gensim} \cite{rehurek_lrec} library. We fine-tuned the number of topics parameters to get the optimal value of TC-NPMI.

Next, we evaluate different components in the BERT-based topic model presented earlier (Figure \ref{fig:framework} from where we have five major approaches:  1.) SBERT-UMAP-HDBSCAN, 2.) SBERT-UMAP-KMeans,    3.) USE-UMAP-HDBSCAN,   4.) USE-UMAP-KMeans, and  5.) LDA. In Table \ref{tab:metric}, we observe that in the case of the Bhagavad Gita,  the
combination of USE-UMAP-KMeans gives the best TC-NPMI score on both the datasets with a very slight difference  when compared to USE-UMAP-HDBSCAN and SBERT-UMAP-KMeans. Note that high TC-NPMI results indicate better results. In the case of the Upanishads, we find a similar trend. We also observe that LDA does not perform   well, even after fine-tuning the number of topics parameters to optimize the topic coherence. 

\begin{figure}
         \centering
         \includegraphics[width=.95\linewidth]{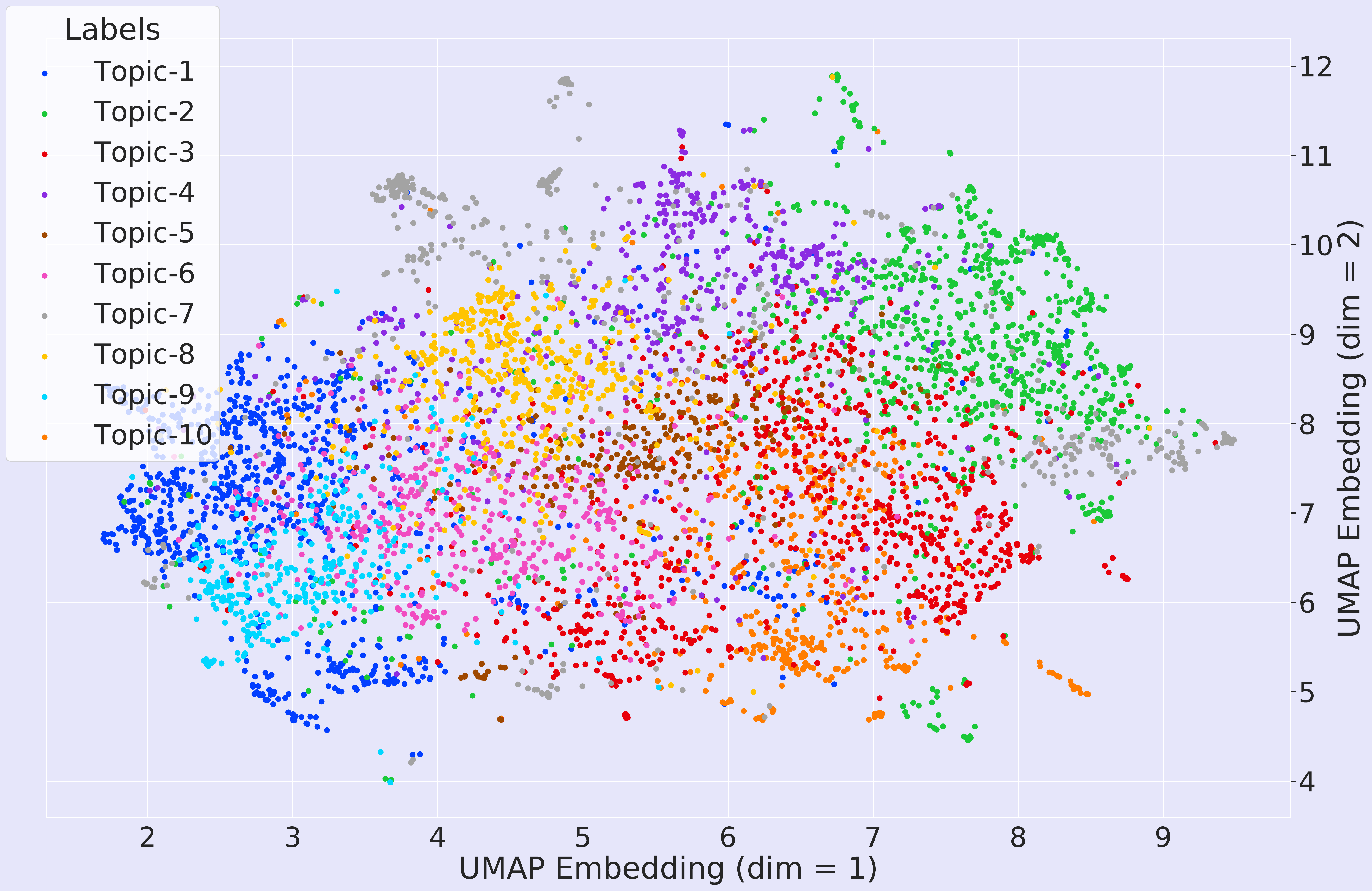}
         \caption{Visualisation of different topics of 108 Upanishads }
         \label{fig:topics-108upan}
\end{figure}

Although the use of  KMeans for the clustering component gives the best result, we choose USE-UMAP-HDBSCAN  to find the topic similarity between the Upanishads and The Bhagavad Gita in the next section.  This is because  HDBSCAN   does not require us to specify the number of clusters, that corresponds to the number of topics, beforehand. USE-UMAP-HDBSCAN gave 18 topics for the corpus - \textit{the Upanishads}\cite{easwaran1987upanishads} for the optimal value of the topic coherence mentioned in Table \ref{tab:metric}.  Similarly, we got 14 topics from the Bhagavad Gita\cite{easwaran1984bhagavad}. In the case of  the 108 Upanishads which contains larger number of documents as compared to the rest of the texts,  we got more topics for the optimal values of topic coherence.  However, we reduced the number of topics using hierarchical topic reduction \cite{angelov2020top2vec} in some of cases for example, while comparing the topic similarity of the Bhagavad Gita and the Upanishads. Since the number of documents and words are different for different corpus as seen from Table \ref{tab:corpus_stat}, the number of topics obtained are different for different corpus. For example, in the Ten Principal Upanishads -- there are 1267 documents and we got 28 topics for them at the optimal value of topic coherence. Similarly for 108 Upanishads, there are  6191 documents which gives 115 topics (Table \ref{tab:metric}) for the model SBERT-UMAP-HDBSCAN at the optimal value of topic coherence.
Also, while plotting the semantic space for the different topics obtained by our model as shown in Figure \ref{fig:eknath_gita_upan_topics}, Figure \ref{fig:topics-108upan}, and Figure \ref{fig:fourgroups108upa}, we reduced the number of topics to 10 in order to visualize the topic's semantic space clearly.

\begin{table*}[!htbp]
\small
\begin{adjustwidth}{-2.25in}{0in}
\begin{tabular}{||c|cccccccc|}
\hline
\multirow{3}{*}{Model} &
  \multicolumn{8}{c|}{Datasets} \\ \cline{2-9} 
 &
  \multicolumn{2}{c|}{Bhagavad Gita} &
  \multicolumn{2}{c|}{The Upanishads} &
  \multicolumn{2}{c|}{Ten Principal Upanishads} &
  \multicolumn{2}{c|}{108 Upanishads} \\ \cline{2-9} 
 &
  \multicolumn{1}{c|}{\# Topics} &
  \multicolumn{1}{c|}{TC-NPMI} &
  \multicolumn{1}{c|}{\# Topics} &
  \multicolumn{1}{c|}{TC-NPMI} &
  \multicolumn{1}{c|}{\# Topics} &
  \multicolumn{1}{c|}{TC-NPMI} &
  \multicolumn{1}{c|}{\# Topics} &
  TC-NPMI \\ \hline\hline
SBERT-UMAP-HDBSCAN &
  \multicolumn{1}{c|}{14} &
  \multicolumn{1}{c|}{0.70} &
  \multicolumn{1}{c|}{18} &
  \multicolumn{1}{c|}{0.67} &
  \multicolumn{1}{c|}{28} &
  \multicolumn{1}{c|}{0.70} &
  \multicolumn{1}{c|}{115} &
  0.63 \\ \hline
SBERT-UMAP-KMeans &
  \multicolumn{1}{c|}{14} &
  \multicolumn{1}{c|}{0.72} &
  \multicolumn{1}{c|}{18} &
  \multicolumn{1}{c|}{0.69} &
  \multicolumn{1}{c|}{28} &
  \multicolumn{1}{c|}{0.73} &
  \multicolumn{1}{c|}{115} &
  0.66 \\ \hline
USE-UMAP-HDBSCAN &
  \multicolumn{1}{c|}{14} &
  \multicolumn{1}{c|}{0.73} &
  \multicolumn{1}{c|}{18} &
  \multicolumn{1}{c|}{0.67} &
  \multicolumn{1}{c|}{32} &
  \multicolumn{1}{c|}{0.73} &
  \multicolumn{1}{c|}{125} &
  0.64 \\ \hline
USE-UMAP-KMeans &
  \multicolumn{1}{c|}{14} &
  \multicolumn{1}{c|}{0.73} &
  \multicolumn{1}{c|}{18} &
  \multicolumn{1}{c|}{0.69} &
  \multicolumn{1}{c|}{32} &
  \multicolumn{1}{c|}{0.71} &
  \multicolumn{1}{c|}{125} &
  0.67 \\ \hline
LDA &
  \multicolumn{1}{c|}{20} &
  \multicolumn{1}{c|}{0.32} &
  \multicolumn{1}{c|}{20} &
  \multicolumn{1}{c|}{0.29} &
  \multicolumn{1}{c|}{24} &
  \multicolumn{1}{c|}{0.31} &
  \multicolumn{1}{c|}{140} &
  0.39 \\ \hline
\end{tabular}
\caption{Value of topic coherence metric (TC-NPMI) for different Corpus}
\label{tab:metric}
\end{adjustwidth} 
\end{table*}


\subsubsection{Topic similarity between the Bhagavad Gita and the Upanishads}
 
There are studies that suggest that the Bhagavad Gita summarizes the key themes of the Upanishads and various other Hindu texts \cite{sargeant2009bhagavad,singh2011sterling,nicholson2010unifying}.
The Bhagavad Gita along with the Upanishads and the Brahma Sutras is known as the \textit{Prasthanatrayi} \cite{meenaview,rao2013brief,lattanzio2020self,mahanti1986concept,nrugham2017suicide}, literally meaning \textit{the three points of departure \cite{meenaview}, or the three sources \cite{lattanzio2020self}}) which makes the three  foundational texts of the Vedanta school of Hindu philosophy \cite{torwesten1991vedanta,radhakrishnan1914vedanta,isayeva1993shankara,singh2011sterling,nicholson2010unifying}. Sargeant et al.\cite{sargeant2009bhagavad} stated that the Bhagavad Gita is the summation of the Vedanta. Nicholson et al.\cite{nicholson2010unifying} and Singh et al.\cite{singh2011sterling}   regarded the Bhagavad Gita to be the key text of  Vedanta.

\begin{figure*}

\begin{adjustwidth}{-2.25in}{0in}
\includegraphics[width=.95\linewidth]{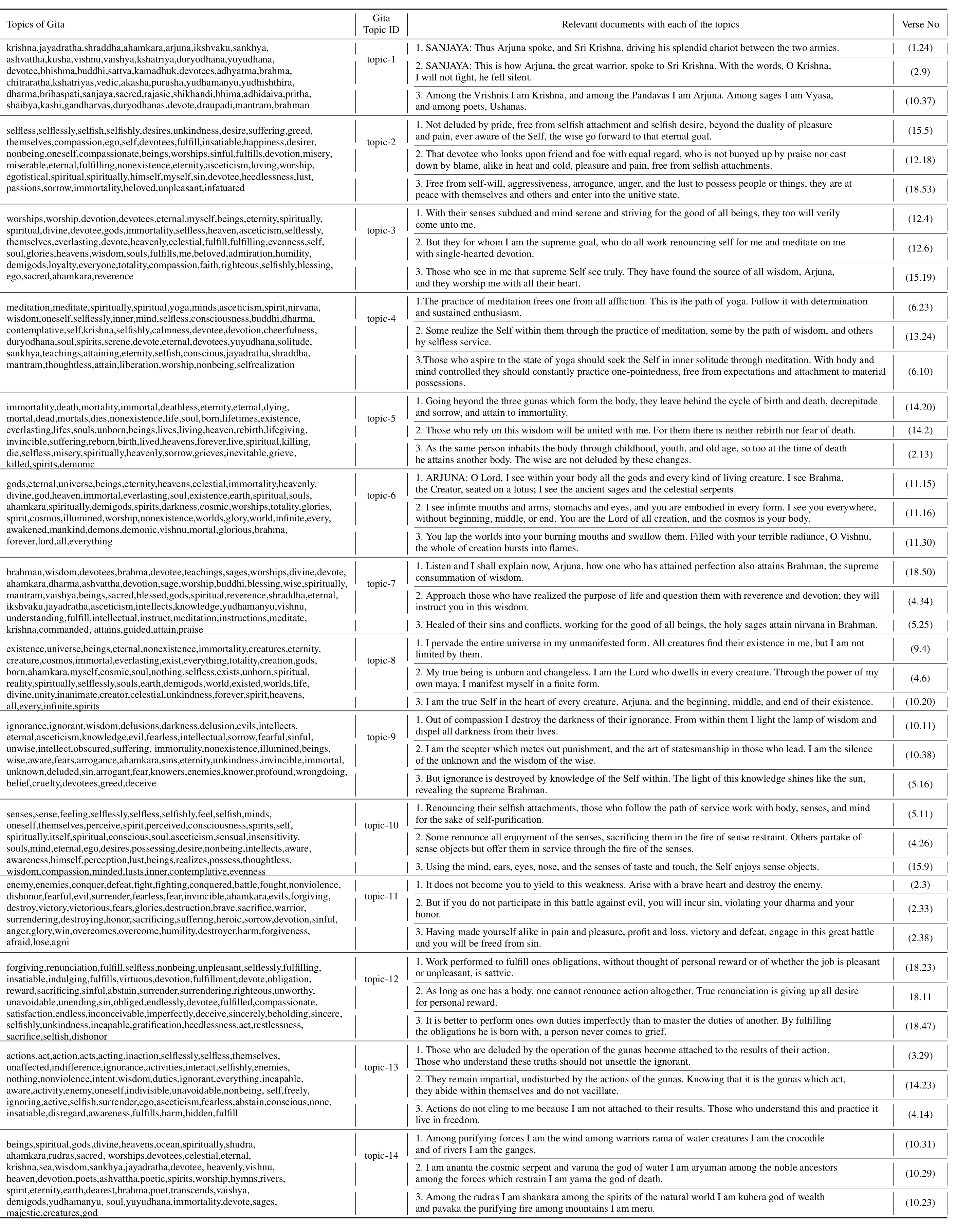}
\caption{Topics of Bhagavad Gita and the most relevant documents(Model: USE-HDBSCAN-UMAP).}
\label{tab:topic-gita-docs}

\end{adjustwidth}
\end{figure*}

\begin{figure*}

\begin{adjustwidth}{-2.25in}{0in}
\includegraphics[width=.99\linewidth]{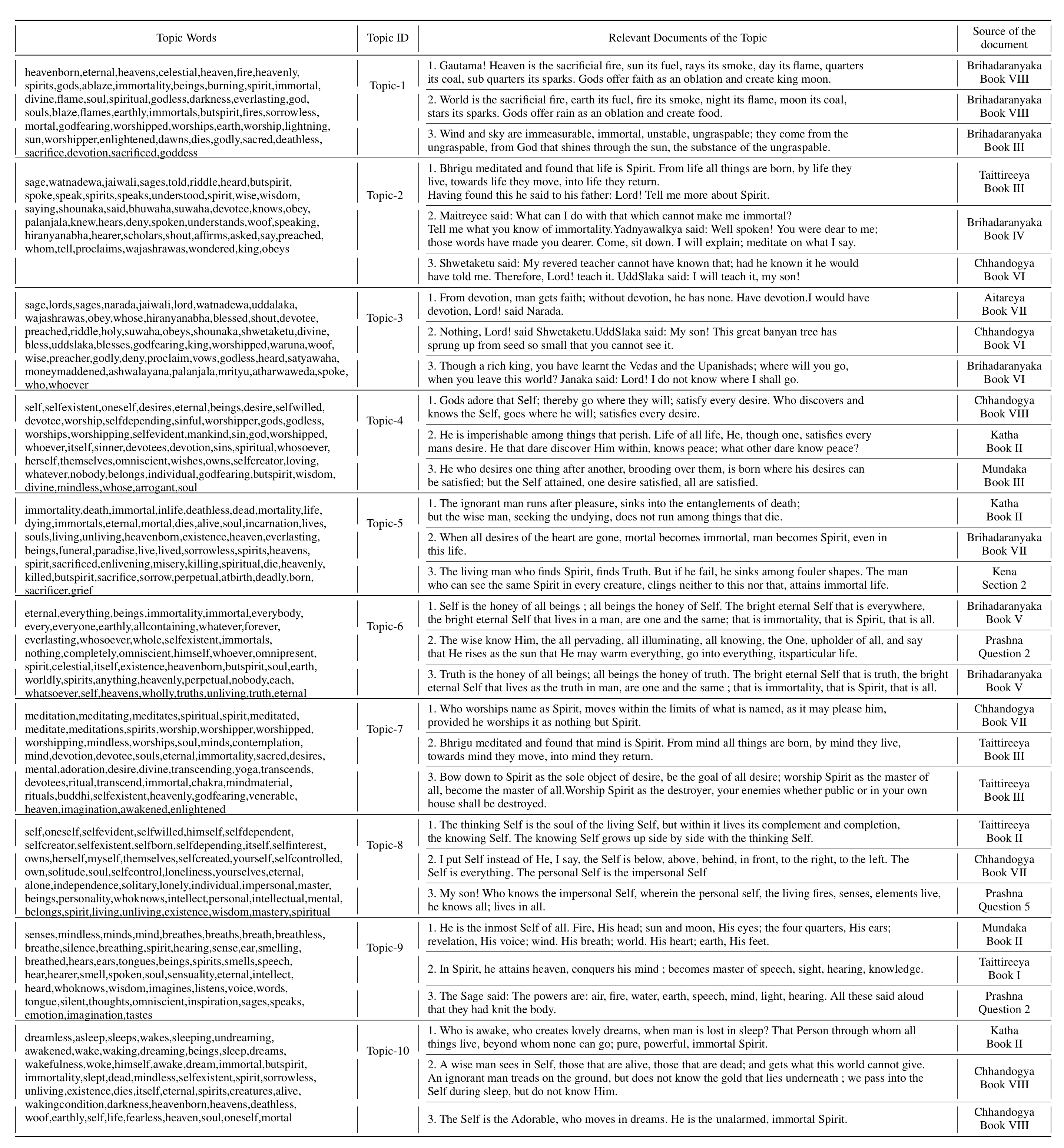}

\caption{Topics of the Ten Principal Upanishads and some of their relevant documents(Model: USE-HDBSCAN-UMAP).}
\label{tab:topic-ten-upan-docs}
\end{adjustwidth}
\end{figure*}

Another source which discusses a direct relationship between the Bhagavad Gita and the Upanishads is the Gita Dhayanam (also sometimes called Gita Dhyana and Dhyana Slokas) which refers to the invocation of the Bhagavad  Gita)\cite{chinmayananda2014srimad,easwaran1984bhagavad,ranganathananda2000universal}.
We need to note that Gita Dhayanam is an accompanying text with 9 verses used for prayer and meditation that complements the Bhagavad Gita. These 9 verses are attributed traditionally to Sri Madhusudana Sarasvati and are generally chanted by the students of Gita before they start their daily studies\cite{chinmayananda2014srimad}. These verses offer salutations to various Hindu entities such as the Vyasa, Lord Krishna, Lord Varuna, Lord Indra, Lord Rudra and the Lord of the Maruta and also characterises the relationship between the Bhagavad Gita and the Upanishads. The 4th verse of the Gita Dhyanam states a direct cow and milk relationship between the Upanishads and the Gita. Eknath Easwaran \cite{easwaran1984bhagavad} translated the 4th verse as \textit{"The Upanishads are the cows milked by Gopala, the son of Nanda, and Arjuna is the calf. Wise and pure men drink the milk, the supreme, immortal nectar of the Gita"}.
Although these   relationships have been studied and retold for centuries, there are no existing studies that establishes a quantitative measure to this relationship using modern language models. Next, we evaluate and discuss similar relationships both quantitatively using a mathematical formulation and also qualitatively by looking at the topics generated by our models as shown in Tables \ref{tab:upan_gita_compare}, \ref{tab:upan-gita-compare-tu}, and Figures \ref{tab:topic-gita-docs}, \ref{tab:topic-ten-upan-docs}.

\begin{figure*}

\begin{adjustwidth}{-2.25in}{0in}
\includegraphics[width=.75\linewidth]{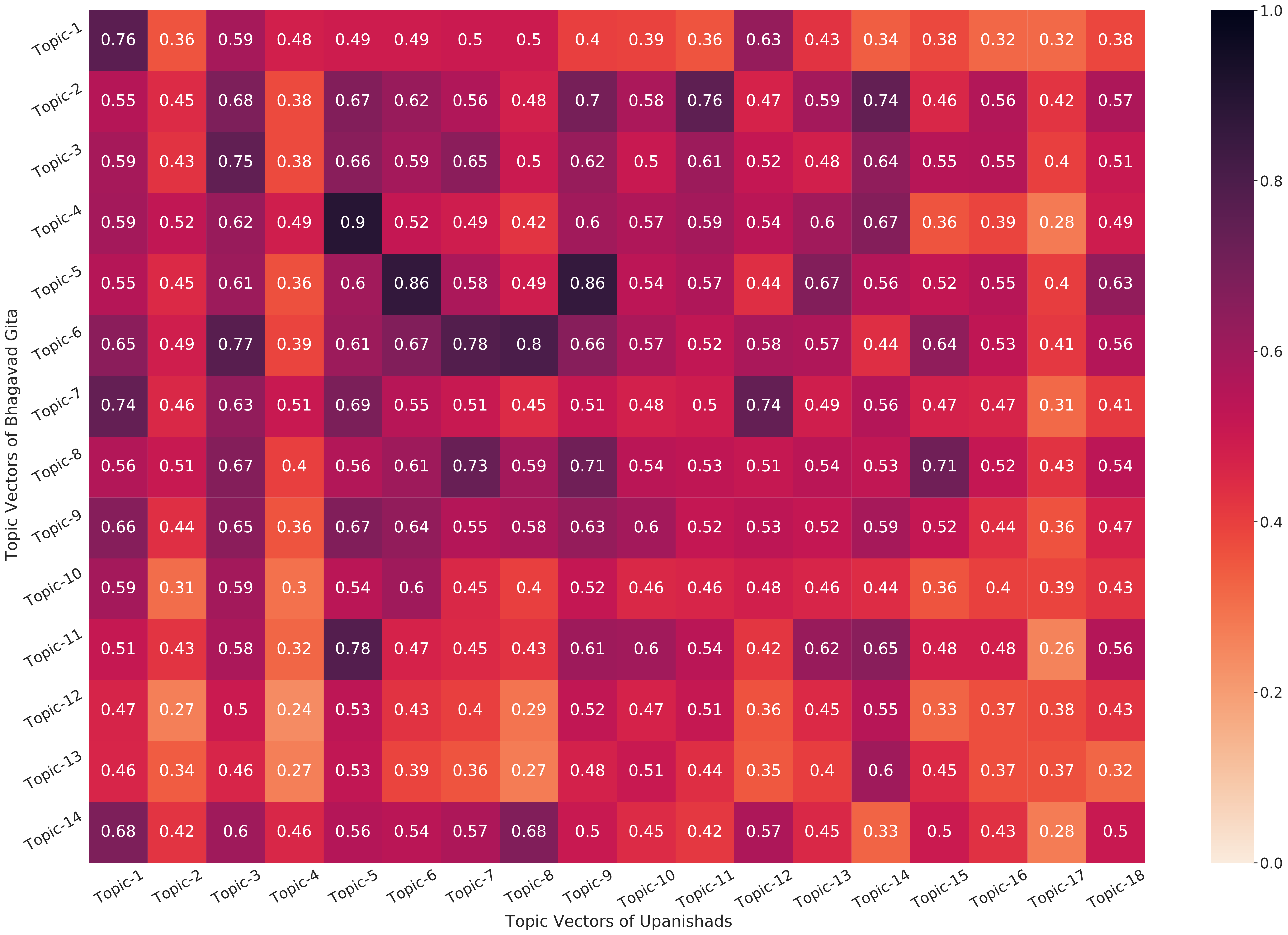}
\caption{Heatmap  showing the similarity between different topics of Bhagavad Gita and Upanishads generated from a selected approach (SBERT-UMAP-HDBSCAN).}
\label{fig:similarity_eknath}

\end{adjustwidth} 
\end{figure*}

In order to evaluate the relationship between the Bhagavad Gita and the Upanishads, we used the obtained topics to find a similarity matrix as shown in the heatmap of Figure \ref{fig:similarity_eknath}. The vertical axis of the heatmap shows the topics of the Bhagavad Gita while the horizontal axis of the heatmap represent the topics of the Upanishads. The heatmap represents the cosine similarity of the topic-vector obtained by the topic model. Therefore, in each  of the topics obtained from the Bhagavad Gita,
we calculate its similarity with all the topics of the Upanishads and then find the topic with maximum similarity. This operation can be mathematically represented by the Equation \ref{eq:best-topic}. We represent the number of topics in Gita by $N_{gita}$ and the number of topics in Upanishads by $N_{upan}$. In each topic $T_i^{gita}$ from the Bhagavad Gita, we explore and find the most similar topic from Upanishads $T_i^{upan}$. The topics and their similarity score can be found in Table \ref{tab:upan_gita_compare} and Table \ref{tab:upan-gita-compare-tu}. We observe a very high similarity in the topics of the Bhagavad Gita and two different texts of Upanishads (shown in Table \ref{tab:upan_gita_compare} and Table \ref{tab:upan-gita-compare-tu}). These tables also show the \textit{mean similarity score} which is given by the average of all the similarity scores as shown in Equation \ref{eq:avg-topic-vec} and given below: 

\begin{subequations}\label{eqn:similarity}
\begin{align}
\operatorname { T_{i}^{upan}}=\mathop{\arg\max}\limits_{{j=1}}\limits^{N_{upan}}\operatorname{Sim}\left(V_{i}^{gita}, V_{j}^{upan}\right)\label{eq:best-topic}\\
\operatorname { AvgSim}=\frac{\sum\limits_{i=1}^{N_{gita}}\mathop{\max}\limits_{{j=1}}\limits^{N_{upan}}\operatorname{Sim}\left(V_{i}^{gita}, V_{j}^{upan}\right)}{N_{gita}}\label{eq:avg-topic-vec}
\end{align}
\end{subequations}

where  $V_i^{gita}$ and $V_i^{upan}$ represent the $i^{th}$ topic vectors of the Bhagavad Gita and the Upanishads, respectively. $Sim(.)$ represent the similarity measure defined by equation \ref{eq:cosine}, which is cosine similarity in our case. There are various other measures of similarity score between two vectors; however, the cosine similarity is used widely in the literature \cite{salicchi2021pihkers,thongtan2019sentiment,gunawan2018implementation}. One of the major reason for this is its interpretability. Value of cosine similarity between any two vector lies between 0 and 1. A value closer to 1 represent that vectors are almost similar to each other and a value closer to 0 represent that they are completely dissimilar.

The cosine similarity between any two vectors $U$ and $V$ is represented by Equation \ref{eq:cosine}. Since the topic vector contains contextual and thematic information about a topic, the similarity score gives us extent of closeness of the themes and topics of the Bhagavad Gita and the Upanishads.
\begin{ceqn}
\begin{align}
   \text{Sim(U, V)}=\cos(\theta )= \frac{\mathbf {U} \cdot \mathbf {V}}{\|\mathbf {U} \|\|\mathbf {V} \|}
   \label{eq:cosine}
\end{align}
\end{ceqn}
We can observe from the Table \ref{tab:upan_gita_compare} that a number of the topics of Bhagavad Gita are similar to the topics of the Upanishads with more than 70\% similarity. We also find that topic 4 of the Bhagavad Gita is similar to that of the topic 5 of the Upanishads (Eknath Easwaran) with a similarity of 90\%. We can see that both of these topics contains almost similar words. Similarly, topic-5 of the Bhagavad Gita has a similarity of 86\% when compared with topic 8 of the Upanishads. Both of these topics are are related to  immortality and death. The similarity can be observed via    Table \ref{tab:upan_gita_compare}; for example, topic-1 of both Bhagavad Gita and the Upanishads (Eknath Easwaran) consists of the words related to Hindu deities and entities such as  Krishna, Arjuna, Vishnu and Samashrava,  they also have a similarity of 76\%.

\begin{figure*}

\begin{adjustwidth}{-2.25in}{0in}
\includegraphics[width=.69\linewidth]{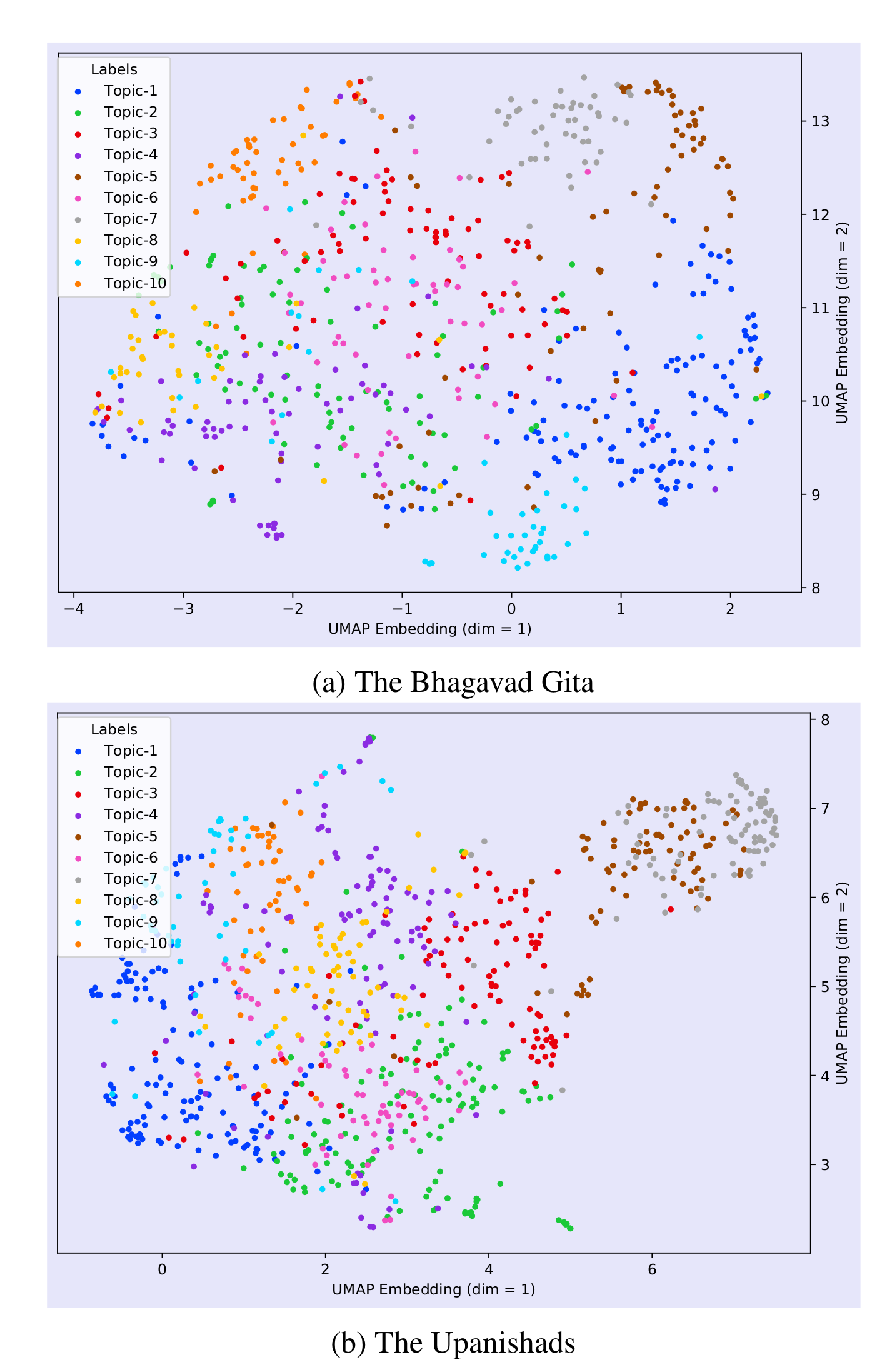}
\caption{Visualization of the semantic space of The Bhagavad Gita and The Upanishads with topic labels.}
\label{fig:eknath_gita_upan_topics}

\end{adjustwidth}
\end{figure*}

Figure \ref{fig:eknath_gita_upan_topics} represents a visualization of the semantic space of the Bhagavad Gita and the Upanishads with given topic labels. Although we find in Table \ref{tab:metric} that Bhagavad Gita and the Upanishads gave 14 and 18 topics respectively, we are only presenting 10 topics from both of these texts in order to have a clear visualization. Each dots in the diagram represent the two dimensional (2D) embedding of each of the documents of the corpus. These topics can be seen in Figure \ref{tab:topic-gita-docs} along with some of the most relevant documents of the text with their source. Figure \ref{tab:topic-gita-docs}  represents the themes related to the deities and the entities of the Hindu philosophy. We can also observe that documents relevant to topic-1 have been originated form chapter 1, 3 and 10. These all are the verses containing the name of the Hindu deities. Topic-2 of the same table encapsulate the idea of self, worship, desire and fulfillment. 
A similar pattern can be observed in Table \ref{tab:upan-gita-compare-tu} which represent the topics and documents of the Ten Principal Upanishads \cite{swami2012ten}.

\begin{figure*}

\begin{adjustwidth}{-2.25in}{0in}
\includegraphics[width=.75\linewidth]{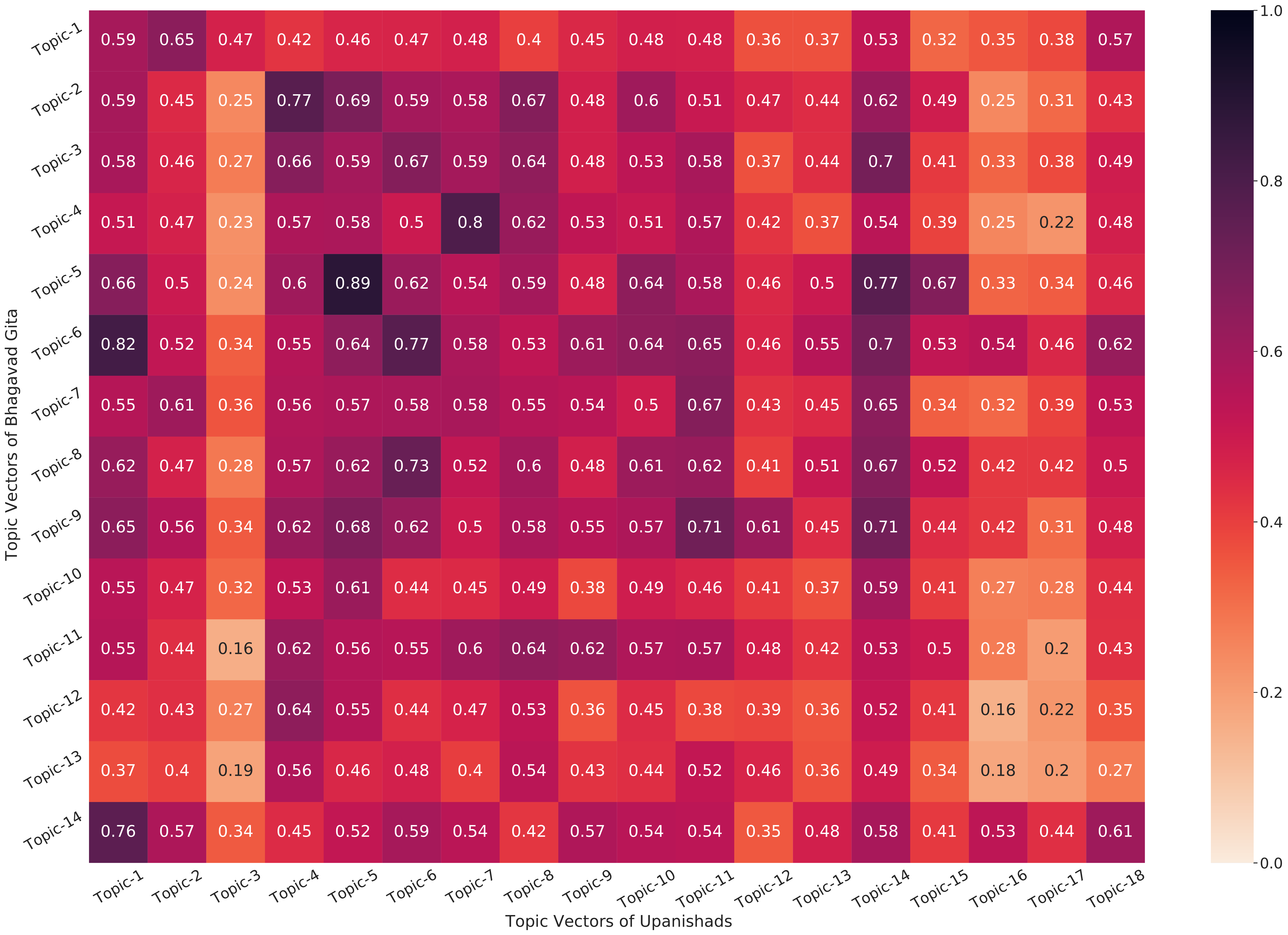}
\caption{Heatmap  showing the similarity between different topics of Bhagavad Gita(Eknath Easwaran) and the Ten Principal Upanishads(Shri Purohit Swami) generated from a selected approach (SBERT-UMAP-HDBSCAN).}
\label{fig:similarity_tu_use}

\end{adjustwidth} 
\end{figure*}

\begin{table*}
\begin{adjustwidth}{-2.25in}{0in}
\scriptsize
\begin{tabular}{|l|c|l|c|l|}
\hline
Topics of Gita &
  \begin{tabular}[c]{@{}c@{}}Gita \\ Topic ID\end{tabular} &
  Most Similar topics in Upanishads &
  \begin{tabular}[c]{@{}c@{}}Upanishads \\ Topic ID\end{tabular} &
  Similarity Score \\ \hline
\begin{tabular}[c]{@{}l@{}}krishna,jayadratha,shraddha,ahamkara,\\ arjuna,ikshvaku,sankhya,ashvattha,kusha,vishnu\end{tabular} &
  topic-1 &
  \begin{tabular}[c]{@{}l@{}}sage,wisdom,devotee,sages,vishnu,mahabharata,\\ devotees,samashrava,hindu,mahavakyas,theravada\end{tabular} &
  topic-1 &
  0.76 \\ \hline
\begin{tabular}[c]{@{}l@{}}selfless,selflessly,selfish,selfishly,desires,\\ unkindness,desire,suffering,greed,themselves\end{tabular} &
  topic-2 &
  \begin{tabular}[c]{@{}l@{}}desires,happiness,eternal,selfless,beings,spiritual,\\ existence,spirituality,desire,joy,eternity,buddhism\end{tabular} &
  topic-11 &
  0.76 \\ \hline
\begin{tabular}[c]{@{}l@{}}worships,worship,devotion,devotees,eternal,\\ myself,beings,eternity,spiritually,spiritual\end{tabular} &
  topic-3 &
  \begin{tabular}[c]{@{}l@{}}eternal,divine,deity,eternity,lords,lord,everlasting,\\ devotional,omnipotent,gods,beings,soul,beloved\end{tabular} &
  topic-3 &
  0.75 \\ \hline
\begin{tabular}[c]{@{}l@{}}meditation,meditate,spiritually,spiritual,\\ yoga,minds,asceticism,spirit,nirvana,wisdom\end{tabular} &
  topic-4 &
  \begin{tabular}[c]{@{}l@{}}meditation,meditating,meditates,meditate,meditated,\\ minds,mind,spiritually,interiorize,enlightenment,spiritual\end{tabular} &
  topic-5 &
  0.90 \\ \hline
\begin{tabular}[c]{@{}l@{}}immortality,death,mortality,immortal,deathless,\\ eternity,eternal,dying,mortal,dead,mortals\end{tabular} &
  topic-5 &
  \begin{tabular}[c]{@{}l@{}}immortality,death,immortal,mortality,deathless,\\ mortal,dying,mortals,eternity,deathlessness,eternal\end{tabular} &
  topic-6 &
  0.86 \\ \hline
\begin{tabular}[c]{@{}l@{}}gods,eternal,universe,beings,eternity,heavens,\\ celestial,immortality,heavenly,divine,god\end{tabular} &
  topic-6 &
  \begin{tabular}[c]{@{}l@{}}celestial,sun,heavens,earth,earthly,heaven,heavenly,\\ luminous,sunrise,sky,universe,illumined,light,illumine\end{tabular} &
  topic-8 &
  0.80 \\ \hline
\begin{tabular}[c]{@{}l@{}}brahman,wisdom,devotees,brahma,\\ devotee,teachings,sages,worships,divine,devote\end{tabular} &
  topic-7 &
  \begin{tabular}[c]{@{}l@{}}sage,wisdom,devotee,sages,vishnu,mahabharata,devotees,\\ samashrava,hindu,mahavakyas,theravada,hindus,buddhi\end{tabular} &
  topic-1 &
  0.74 \\ \hline
\begin{tabular}[c]{@{}l@{}}existence,universe,beings,eternal,nonexistence,\\ immortality,creatures,eternity,creature,cosmos\end{tabular} &
  topic-8 &
  \begin{tabular}[c]{@{}l@{}}universe,omnipotent,eternal,cosmos,eternity,beings,cosmic,\\ immortal,gods,celestial,deity,beyondness,god,heavens\end{tabular} &
  topic-7 &
  0.73 \\ \hline
\begin{tabular}[c]{@{}l@{}}ignorance,ignorant,wisdom,delusions,darkness,\\ delusion,evils,intellects,eternal,asceticism\end{tabular} &
  topic-9 &
  \begin{tabular}[c]{@{}l@{}}meditation,meditating,meditates,meditate,meditated,\\ minds,mind,spiritually,interiorize,enlightenment,spiritual\end{tabular} &
  topic-5 &
  0.67 \\ \hline
\begin{tabular}[c]{@{}l@{}}senses,sense,feeling,selflessly,selfless,selfishly,\\ feel,selfish,minds,oneself,themselves,perceive\end{tabular} &
  topic-10 &
  \begin{tabular}[c]{@{}l@{}}meditation,meditating,meditates,meditate,meditated,\\ minds,mind,spiritually,interiorize,enlightenment,spiritual\end{tabular} &
  topic-5 &
  0.78 \\ \hline
\begin{tabular}[c]{@{}l@{}}enemy,enemies,conquer,defeat,fight,fighting,\\ conquered,battle,fought,nonviolence,dishonor\end{tabular} &
  topic-11 &
  \begin{tabular}[c]{@{}l@{}}immortality,death,immortal,mortality,deathless,mortal,\\ dying,mortals,eternity,deathlessness,eternal,dead,deaths\end{tabular} &
  topic-6 &
  0.60 \\ \hline
\begin{tabular}[c]{@{}l@{}}forgiving,renunciation,fulfill,selfless,nonbeing,\\ unpleasant,selflessly,fulfilling,insatiable,indulging\end{tabular} &
  topic-12 &
  \begin{tabular}[c]{@{}l@{}}selfs,self,selfless,oneself,himself,themselves,selfish,ego,\\ itself,egoism,yourself,independently,ourselves,autonomic\end{tabular} &
  topic-14 &
  0.55 \\ \hline
\begin{tabular}[c]{@{}l@{}}actions,act,action,acts,acting,inaction,selflessly,\\ selfless,themselves,unaffected,indifference,ignorance\end{tabular} &
  topic-13 &
  \begin{tabular}[c]{@{}l@{}}selfs,self,selfless,oneself,himself,themselves,selfish,ego,\\ itself,egoism,yourself,independently,ourselves,autonomic\end{tabular} &
  topic-14 &
  0.60 \\ \hline
\begin{tabular}[c]{@{}l@{}}beings,spiritual,gods,divine,heavens,ocean,\\ spiritually,shudra,ahamkara,rudras,sacred,worships\end{tabular} &
  topic-14 &
  \begin{tabular}[c]{@{}l@{}}sage,wisdom,devotee,sages,vishnu,mahabharata,devotees,\\ samashrava,hindu,mahavakyas,theravada,hindus,buddhi\end{tabular} &
  topic-1 &
  0.68 \\ \hline
   \multicolumn{4}{|c|}{Mean Similarity Score(AvgSim)}  &0.73 \\ \hline
 \end{tabular}
\caption{Topics of the Bhagavad Gita(Eknath Easwaran ) with most similar topics from the Upanishads(Eknath Easwaran)}
\label{tab:upan_gita_compare} 

\end{adjustwidth}
\end{table*}

\begin{table*}[!htb]
\begin{adjustwidth}{-2.25in}{0in}
\scriptsize
\begin{tabular}{|l|c|l|c|c|}
\hline
Topics of Gita &
  \begin{tabular}[c]{@{}c@{}}Gita \\ Topic ID\end{tabular} &
  Most Similar topics in Upanishads &
  \begin{tabular}[c]{@{}c@{}}Upanishads \\ Topic ID\end{tabular} &
  \begin{tabular}[c]{@{}c@{}}Similarity \\ Score\end{tabular} \\ \hline
\begin{tabular}[c]{@{}l@{}}krishna,jayadratha,shraddha,ahamkara,arjuna,\\ ikshvaku,sankhya,ashvattha,kusha,vishnu\end{tabular} &
  topic-1 &
  \begin{tabular}[c]{@{}l@{}}sage,watnadewa,jaiwali,sages,told,riddle,heard,\\ butspirit,spoke,whoknows,thathe,speak,spirits\end{tabular} &
  topic-2 &
  0.65 \\ \hline
\begin{tabular}[c]{@{}l@{}}selfless,selflessly,selfish,selfishly,desires,\\ unkindness,desire,suffering,greed,themselves\end{tabular} &
  topic-2 &
  \begin{tabular}[c]{@{}l@{}}himself,self,selfexistent,oneself,desires,eternal,\\ beings,desire,selfwilled,devotee,worship,sinful\end{tabular} &
  topic-4 &
  0.77 \\ \hline
\begin{tabular}[c]{@{}l@{}}worships,worship,devotion,devotees,eternal,\\ myself,beings,eternity,spiritually,spiritual\end{tabular} &
  topic-3 &
  \begin{tabular}[c]{@{}l@{}}immortality,immortal,immortals,heaven,eternal,\\ heavenborn,heavens,heavenly,celestial,paradise\end{tabular} &
  topic-14 &
  0.70 \\ \hline
\begin{tabular}[c]{@{}l@{}}meditation,meditate,spiritually,spiritual,yoga,\\ minds,asceticism,spirit,nirvana,wisdom\end{tabular} &
  topic-4 &
  \begin{tabular}[c]{@{}l@{}}meditation,meditating,meditates,spiritual,spirit,\\ meditated,meditate,meditations,spirits,butspirit\end{tabular} &
  topic-7 &
  0.80 \\ \hline
\begin{tabular}[c]{@{}l@{}}immortality,death,mortality,immortal,deathless,\\ eternity,eternal,dying,mortal,dead,mortals\end{tabular} &
  topic-5 &
  \begin{tabular}[c]{@{}l@{}}immortality,death,immortal,inlife,deathless,dead,\\ mortality,life,dying,immortals,eternal,mortal,dies\end{tabular} &
  topic-5 &
  0.89 \\ \hline
\begin{tabular}[c]{@{}l@{}}gods,eternal,universe,beings,eternity,heavens,\\ celestial,immortality,heavenly,divine,god\end{tabular} &
  topic-6 &
  \begin{tabular}[c]{@{}l@{}}heavenborn,eternal,heavens,celestial,heaven,fire,heavenly,\\ spirits,gods,ablaze,immortality,beings,spirit,immortal\end{tabular} &
  topic-1 &
  0.82 \\ \hline
\begin{tabular}[c]{@{}l@{}}brahman,wisdom,devotees,brahma,devotee,\\ teachings,sages,worships,divine,devote\end{tabular} &
  topic-7 &
  \begin{tabular}[c]{@{}l@{}}knowledge,whoknows,knowthe,wisdom,known,\\ unknowable,knower,knowing,knew,knows,omniscient\end{tabular} &
  topic-11 &
  0.67 \\ \hline
\begin{tabular}[c]{@{}l@{}}existence,universe,beings,eternal,nonexistence,\\ immortality,creatures,eternity,creature\end{tabular} &
  topic-8 &
  \begin{tabular}[c]{@{}l@{}}eternal,everything,beings,immortality,immortal,everybodys,\\ everybody,all,every,everyone,earthly,allcontaining,whatever\end{tabular} &
  topic-6 &
  0.73 \\ \hline
\begin{tabular}[c]{@{}l@{}}ignorance,ignorant,wisdom,delusions,darkness,\\ delusion,evils,intellects,eternal,asceticism\end{tabular} &
  topic-9 &
  \begin{tabular}[c]{@{}l@{}}knowledge,whoknows,knowthe,wisdom,known,\\ unknowable,knower,knowing,knew,knows,omniscient\end{tabular} &
  topic-11 &
  0.71 \\ \hline
\begin{tabular}[c]{@{}l@{}}enemy,enemies,conquer,defeat,fight,fighting,\\ conquered,battle,fought,nonviolence,dishonor\end{tabular} &
  topic-10 &
  \begin{tabular}[c]{@{}l@{}}immortality,death,immortal,inlife,deathless,dead,\\ mortality,life,dying,immortals,eternal,mortal,dies\end{tabular} &
  topic-5 &
  0.61 \\ \hline
\begin{tabular}[c]{@{}l@{}}senses,sense,feeling,selflessly,selfless,selfishly,\\ feel,selfish,minds,oneself,themselves,perceive\end{tabular} &
  topic-11 &
  \begin{tabular}[c]{@{}l@{}}self,oneself,selfevident,selfwilled,himself,selfdependent,\\ selfcreator,selfexistent,selfborn,selfdepending,itself,selfinterest\end{tabular} &
  topic-8 &
  0.64 \\ \hline
\begin{tabular}[c]{@{}l@{}}forgiving,renunciation,fulfill,selfless,nonbeing,\\ unpleasant,selflessly,fulfilling,insatiable,indulging\end{tabular} &
  topic-12 &
  \begin{tabular}[c]{@{}l@{}}himself,self,selfexistent,oneself,desires,eternal,beings,desire,\\ selfwilled,devotee,worship,sinful,worshipper,gods,godless\end{tabular} &
  topic-4 &
  0.64 \\ \hline
\begin{tabular}[c]{@{}l@{}}actions,act,action,acts,acting,inaction,selflessly,\\ selfless,themselves,unaffected,indifference,ignorance\end{tabular} &
  topic-13 &
  \begin{tabular}[c]{@{}l@{}}himself,self,selfexistent,oneself,desires,eternal,beings,desire,\\ selfwilled,devotee,worship,sinful,worshipper,gods,godless\end{tabular} &
  topic-4 &
  0.56 \\ \hline
\begin{tabular}[c]{@{}l@{}}beings,spiritual,gods,divine,heavens,ocean,spiritually,\\ shudra,ahamkara,rudras,sacred,worships,devotees\end{tabular} &
  topic-14 &
  \begin{tabular}[c]{@{}l@{}}heavenborn,eternal,heavens,celestial,heaven,fire,heavenly,\\ spirits,gods,ablaze,immortality,beings,burning,spirit,immortal\end{tabular} &
  topic-1 &
  0.76 \\ \hline
  \multicolumn{4}{|c|}{Mean Similarity Score(AvgSim)}  &0.71 \\ \hline
\end{tabular}%
\caption{Topics of the Bhagavad Gita(Eknath Easwaran ) with most similar topics from the Ten Principal Upanishads(Shri Purohit Swami \& W.B. Yeats)}
\label{tab:upan-gita-compare-tu}
\end{adjustwidth}
\end{table*}

 


\begin{figure*}

\begin{adjustwidth}{-2.25in}{0in}
\includegraphics[width=.69\linewidth]{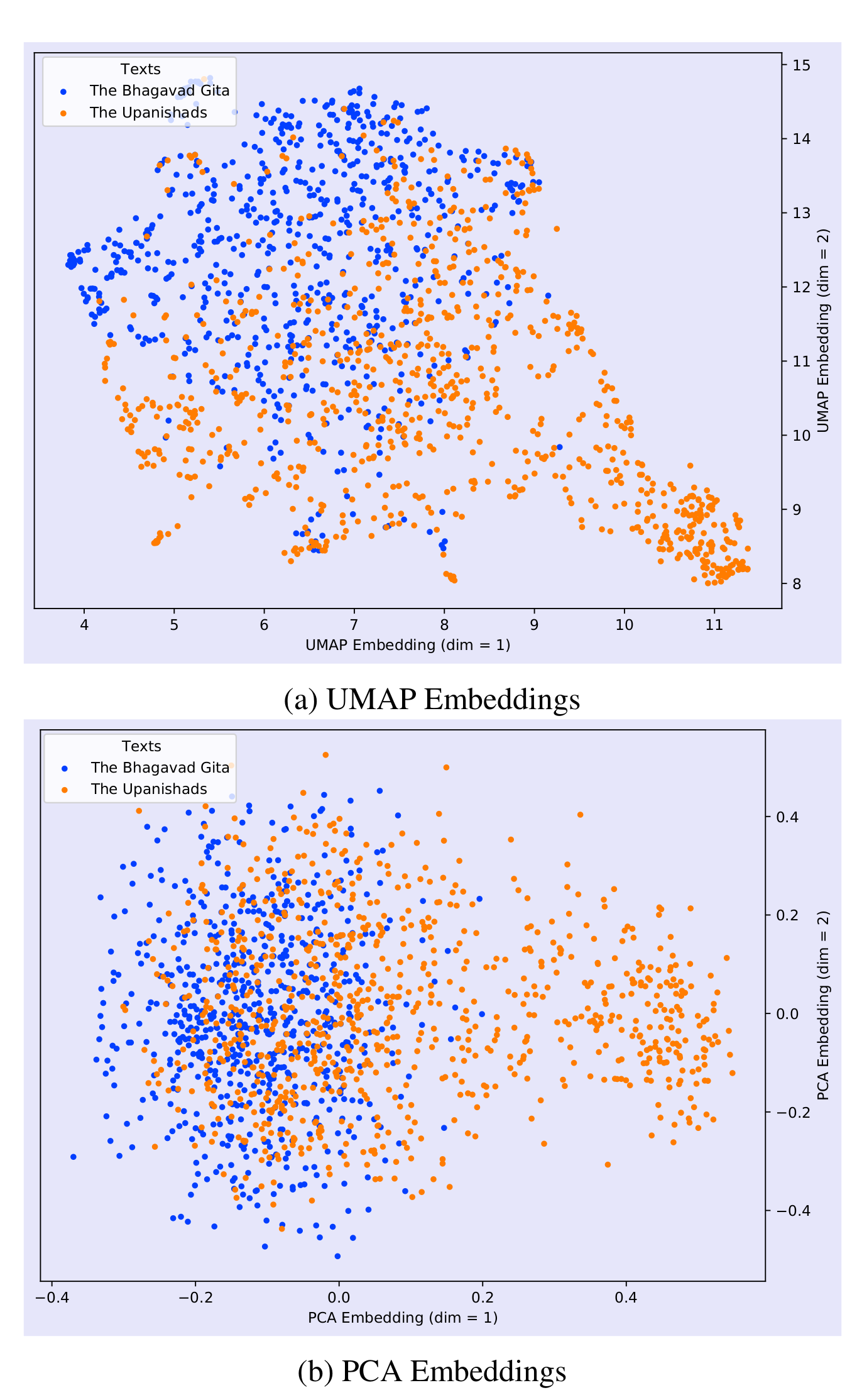}
\caption{Visualization of the Combined Semantic Space of The Bhagavad Gita and The Upanishads (PCA and UMAP).}
\label{fig:gita_in_upan}

\end{adjustwidth}
\end{figure*}

\subsubsection{108 Upanishads}

\begin{table*}[htbp!]
\begin{adjustwidth}{-2.25in}{0in}
\begin{tabular}{|c|c|llllllll|}
\hline
\textbf{Vedas} &
  \textbf{\# Upanishads} &
  \multicolumn{8}{c|}{\textbf{Upanishads}} \\ \hline
Atharva-Veda &
  31 &
  \multicolumn{8}{l|}{\begin{tabular}[c]{@{}l@{}}Prasna,Mundaka,Mandukya, Atahrvasiras,Atharvasikha,Brihajjabala,Devi\\ Nrisimhatapini,Naradaparivrajaka,Sita,Sarabha,Tripadvibhuti-Mahanarayana,\\ Ramarahasya,Ramatapini,Sandilya,Paramahamsaparivrajaka,Annapurna,Surya,\\ Atma,Pasupatabrahmana,Parabrahma,Tripuratapini,Bhavana,Bhasmajabala,\\ Ganapati,Mahavakya,Gopalatapini,Krishna,Hayagriva,Dattatreya,Garuda\end{tabular}} \\ \hline
Krishna-Yajur-Veda &
  32 &
  \multicolumn{8}{l|}{\begin{tabular}[c]{@{}l@{}}Kathavalli,Taittiriyaka,Brahma,Kaivalya,Svetasvatara, Garbha,Varaha,Akshi\\ Narayana,Amritabindu,Amritanada,Kalagnirudra,Kshurika,Sarvasara\\ Sukarahasya,Tejobindu,Dhyanabindu,Brahmavidya,Yogatattva,Dakshinamurti,\\ Skanda,Sariraka,Yogasikha,Ekakshara,Avadhuta,Katharudra,Rudrahridaya,\\ Yoga-kundalini,Panchabrahma,Pranagnihotra,Kalisamtarana,Sarasvatirahasya\end{tabular}} \\ \hline
Sukla-Yajur-Veda &
  19 &
  \multicolumn{8}{l|}{\begin{tabular}[c]{@{}l@{}}Isavasya,Brihadaranyaka,Jabala,Hamsa,Paramahamsa,Paingala,Bhiksu,Tarasara\\ Mantrika,Niralamba,Trisikhibrahmana,Mandalabrahmana,Turiyatita,,Subala\\ Satyayani,Muktika,Advayataraka,Adhyatma,Yajnavalkya\end{tabular}} \\ \hline
Sama Veda &
  16 &
  \multicolumn{8}{l|}{\begin{tabular}[c]{@{}l@{}}Kena,Chandogya,Aruni,Maitrayani,Maitreya,Vajrasuchika,Jabali\\ Rudrakshajabala, Yogachudamani,Vasudeva,Savitri,Darsana\\ Mahat,Sannyasa,Avyakta,Kundika\end{tabular}} \\ \hline
Rig Veda &
  10 &
  \multicolumn{8}{l|}{\begin{tabular}[c]{@{}l@{}}Aitareya,Kaushitakibrahmana,Nadabindu,Atmabodha,Nirvana,Mudgala,\\ Akshamalika ,Tripura,Saubhagyalakshmi, Bahvricha\end{tabular}} \\ \hline
\end{tabular}
\caption{Classification of Upanishads based on original Vedas it is derived from}
\label{tab:vedas_upanishads}
\end{adjustwidth}
\end{table*}
Finally, we apply a selected respective topic modelling approach (USE-UMAP-HDBSCAN) from our topic modelling framework (Figure \ref{fig:framework}) for analysis of the complete 108 Upanishads. We note that the 108 Upanishads are also known as Upanishads that fall under 4 different categories identified by the four Vedas \cite{witzel2003vedas} ( Rig Veda, Samar Veda, Yajur Veda, Artha Veda) which are known as the founding texts of Hinduism. The Rig Veda is the oldest Hindu texts written in ancient Sanskrit   and believed to be remembered orally from guru-student tradition of mantra-recital \cite{yelle2004explaining} thousands of years before being written down \cite{staal2008discovering}. It has been difficult to translated and also understand significance of certain aspects of the Vedas since it has been written in ancient Sanskrit in verse with symbolism \cite{aurobindo2018secret}. The  Upanishads are known as the texts that explain the philosophy of the Vedas and also known as the concluding chapters that have been added to the four Vedas \cite{pandit1974mystic}. Table \ref{tab:vedas_upanishads} gives information about how the 108 Upanishads have been grouped according to their historical  relevance to the respective Vedas.

\begin{figure*}

\begin{adjustwidth}{-2.25in}{0in}
\includegraphics[width=.41\linewidth]{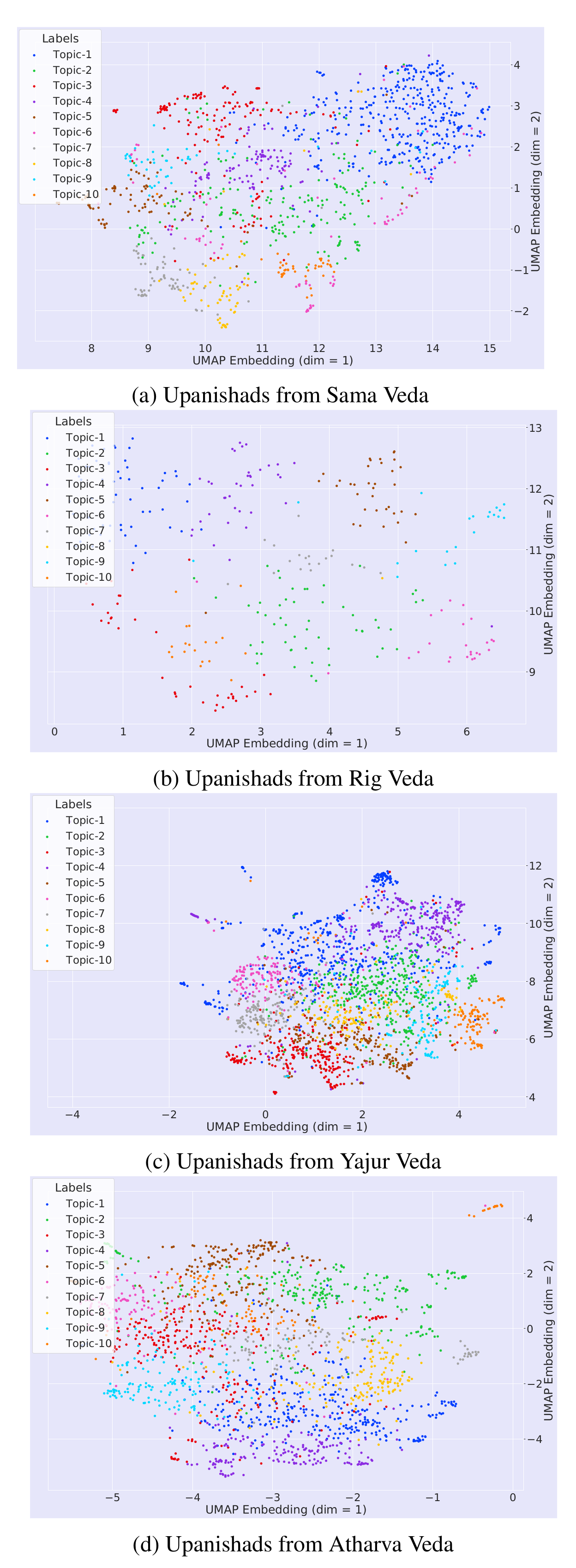}
\caption{Visualization of the semantic space of different parts (based on 4 Vedas) of 108 Upanishads.}
\label{fig:fourgroups108upa}

\end{adjustwidth}
\end{figure*}

 Figure \ref{fig:fourgroups108upa} presents visualization of the semantic space of different parts (divided by 4 Vedas as shown in Table \ref{tab:vedas_upanishads}) of 108 Upanishads.







\section{Discussion}

The high level of semantic and topic similarity between the Bhagavad Gita and the different sets of the Upanishads by the respective authors is not surprising. It verifies well known  thematic similarities as pointed out by Hindu  scholars such as Swami Vivekananda  \cite{vivekananda1937essentials} and western scholars \cite{torwesten1991vedanta}.    Bhagavad Gita is well known as the  central text of Hinduism   that summaries the rest of the Vedic corpus. The Bhagavad Gita is a conversation between Lord Krishna and Arjuna in a  situation where Arjuna has to go to war. The Bhagavad Gita is a chapter from the Mahabharata that uses a conflicting event to summarize philosophy of the Upanishads and the Vedic corpus. The Mahabharata is one of the oldest and longest texts written in verse form in  Sanskrit which describes a historical event    (118,087 sentences, 2,858,609 words)   \cite{das2016computational}. We note that most of Hindu ancient texts have been written in verse so that it can be sung and remembered through an oral tradition in an absence of a writing system.

The goal of Lord Krishna was to motivate Arjuna to do his duty (karma) and go to war to protect ethical standards (dharma) in the society. Krishna, in the Bhagavad Gita begins by renouncing his duties as a warrior. We note that the Mahabharata war, is believed to take place after the Vedas and the Upanishads were composed. Note that by composition, it does not mean that these texts were written, they were sung and verses became key mantras that were remembered through a guru-student tradition for thousands of years.   There are accounts where the Vedas have been mentioned in the Mahabharata. Hence, Krishna is known as a student of the Vedic corpus which also refers to the entire library of Hindu science, literature, history and philosophy. Therefore, the topics in the Upanishads were well known by Lord Krishna and he may have merely used some of the themes to highlight about themes of duty, ethics (dharma) and work (karma) in order to motivate Arjuna to do his duty at the time of need, otherwise, his side (Pandava) would lose the war to the opposition (Kaurava). The Mahabharata war has blood relives on opposing sides of the war battleground known as Kurushetra and hence it was difficult for Arjuna to make a decision either to fight for dharma or renounce his duties and become  a yogi (mystic).   

 
Table \ref{tab:upan_gita_compare} further  compares the topics of the Bhagavad Gita with the Upanishads. We can observe that each of the topic encapsulate some of the ideas expressed in selected verses shown in Tables 7 and 8. If a topic of the Gita and the Upanishads have very high similarity, this represents the fact that the ideas encapsulated by the topics of the Gita and the Upanishads are almost same. In Table \ref{tab:upan_gita_compare}, we can observe that topic 4 of the Bhagavad Gita and topic-5 of the Upanishads have a similarity of $90\%$, this can be seen from the topics also they are representing the similar themes that are related to the ideas of meditation, yoga and spirituality. Similarly, we observe that topic-5 of Gita have a similarity score of $86\%$ when compared  with topic-6 of the Upanishads. Here,  we can also observe  that both topics encapsulate similar ideas of death, mortality and immortality. Similar ideas can be observed in Table \ref{tab:upan-gita-compare-tu} as well the topics of the Bhagavad Gita is compared with the topics of the Upanishads.

Figure \ref{fig:eknath_gita_upan_topics} depicts a representation of the semantic space of the Bhagavad Gita and the Upanishads with topic labels. It represents the lower dimensional embedding of the very high dimensional document vectors. In Figure \ref{fig:eknath_gita_upan_topics}, we  represented only 10 topics in order to retain the clarity of the diagram. Figure \ref{fig:gita_in_upan} shows the UMAP and PCA embedding of the entire document.
In order to generate this plot, we first created the embeddings of each documents and then reduced the embedding to 2D by using PCA  and UMAP. After reducing the dimension, we assigned the labels (\textit{Gita}, and the \textit{Upanishads}) based on the corpus. Figure \ref{fig:gita_in_upan} shows that low-dimensional embeddings reveals very clear overlaps across the documents.

 Even with the presence of translation bias by considering two different translations of the Upanishads, our results demonstrate a very high resemblance between the topics of these two texts, with a mean cosine similarity of more than $70\%$ between the topics of the Bhagavad Gita and those of the Ten Principal Upanishads by Shi Purohit Swami and W.B Yeats. Eight of the fourteen topics extracted from the Bhagavad Gita have a cosine similarity of more than $70\%$ with the topics in the 10 Principal Upanishads, which can also be seen in the table \ref{tab:upan-gita-compare-tu}, and 3 of them have a similarity of more than $80\%$. When considering the translation of both texts by same author as in the case of the Bhagavad Gita\cite{easwaran2007bhagavad} and the Upanishads\cite{easwaran1987upanishads}, we see that average similarity increase to $73\%$ with 9 out of 14 topics having more than $70\%$ similarity and 3 of them having a similarity of more than $80\%$. We also found that topics generated by the BERT based models show very high coherence as compared to that of the LDA. Our best performing model gives a coherence score of $73\%$ on The Bhagavad Gita\cite{easwaran2007bhagavad}, $69\%$ on The Upanishads\cite{easwaran1987upanishads}, $73\%$ on The ten Principal Upanishads\cite{swami2012ten} and $66\%$ on the 108 Upanishads.

 The major limitation is due to the translation bias, which is not present when we take the same translator - this is why we chose  the Upanishads and Bhagavad Gita by Eknath Easwaren in order to limit the bias. However, if we consider the complete 108 Upanishads, such translation bias remain. Moreover, the style and language of the translations not only depend on the translator but on the era. In the case of the 108 Upanishads, a group of translators have contributed which creates further biases. However, in terms of topics uncovered, we find a consistent set of topics that well alight with the respective texts, after manually verifying it.

Further extension can be done by taking the other translations into consideration. \textit{The Ten Principal Upanishads}\cite{swami2012ten} published in 1938, was translated by the Irish poet William Butler Yeats and  Hindu guru Shri Purohit Swami. The translation process occurred between the two authors throughout the 1930s and this book can has been claimed as one of the final works of William Butler Yeats \cite{yeats}. We note that Shri Purohit Swami has also translated the Bhagavad Gita, hence this would be a good companion with Eknath Eashwaren for the respective texts. These extensions could help in refining the proposed framework.

Moreover, in terms of the mythological texts and epics, there are various texts such as the Vishnu Purana, Shiv Purana out of the 18 different Puranas that have underlying topics that are similar. In this study, we focused on philosophical texts, while in future studies, there can be scope for topic modelling from selected texts in the Puranas. The framework can also be used to study texts from other religions, along with n non-religious and non-philosophical  texts. Furthermore, it can be used to study themes of modern poetry, writers, songs and be used to compare different religions and time frames, i.e how the themes changes over different centuries, prior to or after a war or a pandemic (such as the COVID-19).

We note that there exists specialised BERT pre-trained models such as those for medicine  and law \cite{lee2020biobert,tai2020exbert,beltagy2019scibert,chalkidis2020legal,rasmy2021med,muller2020covid}, but there is nothing yet developed  for philosophy. Hindu philosophy is distinct and has terms and ideas that are not present in other philosophical areas (such as western philosophy). Hence, we need specialised pre-trained BERT model for Hindu philosophy which can provide better predictions in related language tasks since it will have better knowledge-base. This work can further be improved using language models for the native Sanskrit text. We intend to explore topic models after building BERT-like language models for Indian philosophical literature written in Sanskrit.

We note that our previous work focused on semantic and sentiment analysis of the Bhagavad Gita translations \cite{chandra2022semantic}. Augmenting semantic and sentiment analysis to our proposed topic modelling framework can provide more insights to the meaning behind the philosophical verses. We plan to build our models in a similar fashion and investigate their variations for texts in three different languages: Hindi, English, and Sanskrit. Finally, post verification study is needed where Sanskrit expert and Hindu philosophers can study the topics uncovered by the proposed framework. 

\textcolor{black}{The Bhagavad Gita and the Upanishads are considerably large texts in the content of religious and philosophical texts. However, the proposed framework can be used for larger corpus such as modelling overlapping topics around the Mahabharata and the Puranas, which are texts that are magnitudes larger than the ones considered in this study. However, we note that the Bhagavad Gita and Upanishads, although smaller  in size are more condensed in philosophy while the Mahabharata is an epic poem.}

\textcolor{black}{In future work, there can be a detailed study of the topics uncovered with a discussion of related texts in Vedic studies that relate to morphology, lexicography, grammar (patterns in sentences), meter (lengthy sentences), and phonology (sound system), etc. 
Furthermore, we need to create processed benchmark text datasets for Indian languages that can benefit NLP applications associated with Indian languages. }


\section{Conclusion and Future work}

We presented a topic modeling framework for Hindu philosophy using state-of-art  deep-learning based models. The use of such technique for studying Hindu texts is relatively novel; however, computational and statistical approaches have been used in the past. The major goal of the study was to link the topics from the Upanishads with the Bhagavad Gita.

 The representation of the low-dimensional embeddings presented in this work reveals a lot of overlap between the Upanishads and the Bhagavad Gita's topics, which adds to our objective of demonstrating the Bhagavad Gita's relationship with the Upanishads. Given the importance of religious literature to a community, employing computational models to verify any of its old and traditional philosophical principles demonstrates the scientific nature of the literature and religion. Despite the fact that the idea of the Gita being the essential extract of the Upanishads has been written and researched in ancient Indian philosophical literature for generations, no attempt has ever been made to substantiate this facts using computational and scientific methodologies. Our research presents a novel way for applying modern deep learning-based methods to a centuries-old philosophical narratives. 
 
\section*{Data and Code}

Python-based open source code and data can be found here
\footnote{\url{https://github.com/sydney-machine-learning/topicmodelling_vedictexts}}.

\section*{Author contributions statement}

R. Chandra devised the project with the main conceptual ideas and   and contributed to overall writing,  literature review and discussion of results.  M. Ranjan provided implementation and experimentation and further contributed in results visualisation and analysis along with writing.

\section*{Acknowledgement}

We thank Shweta Bindal from Indian Institute of Technology - Guwathi   for contributing to discussions about the workflow used in this work.

\section*{Appendix}

\bibliographystyle{plos2015}
\bibliography{language,usyd,sample,sample_,2020June,covid,Hinduism}

\end{document}